\pdfoutput=1
\documentclass[acmtog, review=false, anonymous=false, nonacm=True]{acmart}
\usepackage{blindtext}
\usepackage{graphicx}
\usepackage{algorithm}
\usepackage{algpseudocode}
\usepackage{adjustbox}
\usepackage[english]{babel}
\usepackage{stfloats}
\usepackage{textcomp}
\usepackage{pgfplots}
\usepackage[mathscr]{euscript}
\graphicspath{ {./figures/} }

\AtBeginDocument{%
  \providecommand\BibTeX{{%
    \normalfont B\kern-0.5em{\scshape i\kern-0.25em b}\kern-0.8em\TeX}}}

\setcopyright{acmcopyright}
\copyrightyear{2022}
\acmYear{2022}
\acmDOI{XXXXXXX.XXXXXXX}

\acmJournal{TOG}
\acmVolume{37}
\acmNumber{4}
\acmArticle{111}
\acmMonth{10}

\usepackage{color}
\usepackage{bm}
\usepackage{soul}

\usepackage{caption}
\usepackage{subcaption}
\newif\ifsubmit
\definecolor{changes_color}{rgb}{0.05,0.5,0.3}
\definecolor{mathbrace_color}{rgb}{0.2,0.5,1.0}
\definecolor{light_gray}{rgb}{0.6,0.6,0.6}
\iftrue

      {}

\newenvironment{namedparagraph}[1]{
    \vspace{0.1cm}
    \noindent\textbf{#1}:
        }{}

\newcommand{\beq}{\begin{equation}}
\newcommand{\eeq}{\end{equation}}

\DeclareMathOperator*{\argmax}{argmax}

\newcommand{\argmaxover}[1]{\underset{#1}{\argmax}}

\newcommand{\bracetext}[1]{\text{\tiny #1}}

\newcommand{\coverbrace}[2]{\color{mathbrace_color}\overbrace{\color{black}#1}^{\bracetext{#2}}\color{black}}
\newcommand{\coverbracelines}[3]{\color{mathbrace_color}\overbrace{\color{black}#1}^{\substack{\text{\bracetext{#2}}\\\text{\bracetext{#3}}}}\color{black}}

\newcommand{\AlphaVal}[1]{\mathbf{\alpha}#1}

\newcommand{\ImageVal}[1]{v#1}
\newcommand{\Background}[1]{B#1}

\newcommand{\Foreground}[1]{F#1}

\newcommand{\Video}[1]{{\mathbf{V}}_{#1}}
\newcommand{\ForegroundVideo}{\mathbf{\Foreground{}}}
\newcommand{\BackgroundVideo}{\mathbf{\Background{}}}
\newcommand{\AlphaVideo}{\mathbf{A}}

\newcommand{\Discr}[1]{\mathbf{D}_{#1}}

\newcommand{\Structured}{marginal\xspace}
\newcommand{\Unstructured}{conditional\xspace}

\newcommand{\Layer}[1]{\mathbf{L}_{#1}}
\newcommand{\LayerColor}[1]{\mathbf{C}_{#1}}
\newcommand{\LayerAlpha}[1]{\mathbf{A}_{#1}}
\newcommand{\LayerFlow}[1]{\mathbf{F}_{#1}}
\newcommand{\LayerGTFlow}[1]{\mathbf{F^*}{#1}}
\newcommand{\FactorMatte}[1]{\mathbf{\mathcal{F}}_{#1}}

\newcommand{\CausalMatte}{FactorMatte\xspace}
\newcommand{\OurMethod}{FactorMatte\xspace}
\newcommand{\OmniMatte}{Omnimatte\xspace}

\newcommand{\FL}{foreground layer\xspace}
\newcommand{\RL}{residual layer\xspace}
\newcommand{\BL}{environment layer\xspace}
\newcommand{\posex}{positive examples\xspace}
\newcommand{\Posex}{Positive examples\xspace}

\newcommand{\AlphaF}[1]{\AlphaVal{_f}}
\newcommand{\AlphaB}[1]{\AlphaVal{_b}}
\newcommand{\AlphaR}[1]{\AlphaVal{_r}}

\newcommand{\tForeground}[1]{$\Foreground{#1}$}
\newcommand{\tBackground}[1]{$\Background{#1}$}
\newcommand{\tImageVal}[1]{$\ImageVal{#1}$}

\newcommand{\tLest}{$L_\textit{flow-est}$\xspace}
\newcommand{\tLrecon}{$L_\textit{recon}$\xspace}
\newcommand{\tLCwarp}{$L_\textit{RGB-warp}$\xspace}
\newcommand{\tLAwarp}{$L_{\alpha\textit{-warp}}$\xspace}
\newcommand{\tLsparsity}{$L_{\alpha\textit{-reg}}$\xspace}
\newcommand{\tLadv}{$L_\textit{adv}$\xspace}
\newcommand{\tLmask}{$L_\textit{mask-init}$\xspace}

\newcommand{\tdecompN}{$\mathbf{N}_D$\xspace}

\newcommand{\Lest}{L_\textit{flow-est}\xspace}
\newcommand{\Lrecon}{L_\textit{recon}\xspace}
\newcommand{\LCwarp}{L_\textit{RGB-warp}\xspace}
\newcommand{\LAwarp}{L_{\alpha\textit{-warp}}\xspace}
\newcommand{\Lsparsity}{L_{\alpha\textit{-reg}}\xspace}
\newcommand{\Ladv}{L_\textit{adv}\xspace}
\newcommand{\Lmask}{L_\textit{mask-init}\xspace}

\newcommand{\LogL}[1]{\mathscr{L}_{#1}}

\newcommand{\interactionprior}{conditional mapping}
\newcommand{\InteractionPriors}{Conditional Mapping}
\newcommand{\conditionalmapping}{\interactionprior}
\begin{document}

\newcommand{\PaperTitle}{FactorMatte: Redefining Video Matting for Re-Composition Tasks}

\newcommand{\ShortPaperTitle}{\PaperTitle{}}

\title[\ShortPaperTitle{}]{\PaperTitle{}}

\author{Zeqi Gu}
\affiliation{%
  \institution{Cornell Tech}
  \city{New York, NY}
  \country{USA}}
\email{zg45@cornell.edu}

\author{Wenqi Xian}
\affiliation{%
  \institution{Cornell Tech}
  \city{New York, NY}
  \country{USA}}
\email{wx97@cornell.edu}

\author{Noah Snavely}
\affiliation{%
  \institution{Cornell Tech}
  \city{New York, NY}
  \country{USA}}
\email{snavely@cs.cornell.edu}

\author{Abe Davis}
\affiliation{%
  \institution{Cornell University}
  \city{Ithaca, NY}
  \country{USA}}
\email{abedavis@cornell.edu}

\renewcommand{\shortauthors}{Gu et al}

\begin{abstract}
We propose \emph{factor matting}, an alternative formulation of the video matting problem in terms of counterfactual video synthesis that is better suited for re-composition tasks. The goal of factor matting is to separate the contents of video into independent components, each visualizing a counterfactual version of the scene where contents of other components have been removed. We show that factor matting maps well to a more general Bayesian framing of the matting problem that accounts for complex conditional interactions between layers. Based on this observation, we present a method for solving the factor matting problem that produces useful decompositions even for video with complex cross-layer interactions like splashes, shadows, and reflections. Our method is trained per-video and requires neither pre-training on external large datasets, nor knowledge about the 3D structure of the scene. 
We conduct extensive experiments, and show that our method not only can disentangle scenes with complex interactions, but also outperforms top methods on existing tasks such as classical video matting and background subtraction.
In addition, we demonstrate the benefits of our approach on a range of downstream tasks.

\end{abstract}

\ccsdesc[500]{Computing methodologies~Computer vision representations}
\ccsdesc[300]{Computing methodologies~Rendering}
\ccsdesc[300]{Computing methodologies~Neural networks}

\keywords{matting, video matting, compositing, video layer decomposition}

\begin{teaserfigure}
\includegraphics[width=1\linewidth]{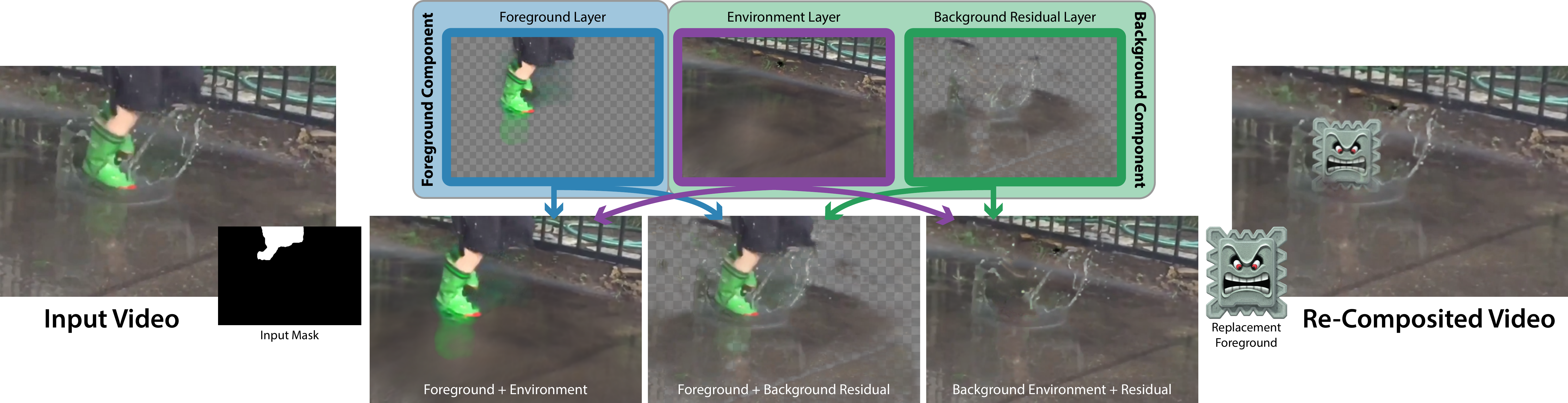}
\caption{
We show that by re-framing the matting problem as counterfactual video synthesis we can produce useful decompositions of video with complex cross-component interactions. Here, the input video (left) shows a child running through a puddle to create splashes. Our method factors the video into two counterfactual components: one for the foot in the foreground (single layer, top left) and another for the background, which includes static parts of the environment (top middle) as well as the splash (top right). The bottom row shows three re-combinations of the component layers, and the image on the right shows a re-composition where the foreground layer has been replaced with a virtual object. To our knowledge, no previous matting methods can handle this kind of complex cross-component interactions.
}
\label{fig:teaser}
\end{teaserfigure}

\maketitle

\section{Introduction}
\label{sec:intro_v2}
The origins of video matting date back to the late 1800's---long before the invention of digital video---with the term ``matting" itself at first referring to the application of matte paint to glass plates that were set in front of a camera during filming.
Once painted, the matte portion of a plate would block parts of the frame from being exposed. The filmmaker could then expose the same film multiple times, using different plates to fill in different parts of the scene with different exposures.
More than a century later, modern digital matting functions much the same way, digitally mixing images according to the compositing equation:
\beq
\ImageVal{_p}=\AlphaVal{_p}\Foreground{_p} + (1-\AlphaVal{_p})\Background{_p}
\label{eq:classicalmatting}
\eeq
which defines an output image value \tImageVal{_p} at image location $p$ as a linear combination of foreground color, \tForeground{_p}, and background color, \tBackground{_p}.
This equation closely mimics the behavior of its analog predecessor, with $\AlphaVal{_p}$ representing the transparency of a glass plate used to film the foreground, and $(1-\AlphaVal{_p})$ representing the transparency of a plate used to film the background. Compared to glass plates, digital mattes are simple to manipulate and need not remain static over time, which makes it easy to re-compose scenes by transforming and re-combining matted content from different sources.
This type of re-compositing can be very powerful under the right conditions, but has some significant limitations as well.
Equation \ref{eq:classicalmatting} assumes that the content of a scene can be factored unambiguously into foreground and background layers, and re-compositing with these layers assumes that each can be manipulated independently of the other. These assumptions place significant constraints on the source material being composited, which
is especially problematic in the context of natural video matting, where the goal is to extract this source material from regular video for later re-compositing.

In this paper, we propose \emph{factor matting}, an adjusted framing of the matting problem that factors video into more independent components for downstream editing tasks.
We formulate factor matting as counterfactual video synthesis. The goal is to factor input video into distinct components, each representing a counterfactual version of the scene with different content removed. We relate each of these counterfactual components to a conditional prior on the appearance of different scene content and show that factor matting closely follows a Bayesian formulation of the matting problem where common limiting assumptions about the independence of different layers have been removed. We then present \emph{\OurMethod{}}, a solution to the factor matting problem that offers a convenient framework for combining classical matting priors with conditional ones based on expected deformations in a scene.
We show that even without training on external data, \OurMethod{}
can yield improved results on traditional matting inputs as well as useful decompositions for inputs with complex foreground-background interactions that have not been addressed in previous work.

\section{Related Work}
\label{sec:relatedwork}
\subsection{Matting Using Priors}
The traditional definition of matting involves inverting Equation \ref{eq:classicalmatting} to recover per-pixel alpha values. As this problem is under-constrained, solving it requires some sort of prior, and it can be useful to interpret previous approaches in terms of the priors they employ. 
\subsubsection{Color-Based Priors}
Priors based solely on color have the advantage of being applicable to static image matting. One of the earliest approaches of this type is color keying, which adopts the prior that the background will be a particular key color~\cite{bluescreenmatting96}. This strategy is effective when one can control the background, for example by filming in front of a green screen. Hence, color keying is still commonly used by filmmakers today.

Scenarios where the background cannot be controlled are often referred to as \emph{natural image matting}. Ruzon and Tomasi were among the pioneers that tackled this problem~\cite{alphaEstimationNaturalImages00}. Their approach was the first to ask a user to partition images into three regions: unambiguous foreground pixels, unambiguous background pixels, and pixels that are an unknown mixture of these two layers. This segmentation, often referred to as a \emph{trimap}, provides a labeled set of representative samples from the foreground and background with which to train data-driven priors. Chuang et al.\ built on this approach by introducing a general Bayesian framework for such problems~\cite{chuang2001bayesian}. Under this framework, matting is performed by solving for a maximum a poteriori (MAP) estimate of the alpha value at each pixel in the uncertain region of the provided trimap. While our method is quite different from these early approaches, we build on their Bayesian reasoning to motivate our approach (Section \ref{sec:classicalmatting}).

Later image matting approaches focused on operating with more relaxed input labels~\cite{iterativestrokematte05}, or optimizing according to detail representations including an image's gradients~\cite{poissonmatting04}, or Laplacian~\cite{closedform08}.

\subsubsection{Motion-Based Priors}
The simplest motion-based prior is basic background subtraction, which assumes that the background content in a scene does not change over time ~\cite{qian1999video, barnich2010vibe}. 
Other early video matting methods used optical flow to interpolate and propagate trimaps across different frames of video (e.g.~\cite{chuang2002video}). 
Subsequent works made more explicit use of temporal coherence by looking at video matting as an operation over a space-time volume of pixels~\cite{bai2009video, lee2010temporally, bai2011towards, choi2012video, li2013motion}.
\cite{lim2018foreground, lim2020learning, tezcan2020bsuv} used background subtraction in their solutions with neural networks. Recent background matting methods~\cite{sengupta2020background, lin2021real} achieved real-time, high-resolution matting but still require a clean background reference image.
Most recently, \cite{lu2021omnimatte} use a variant of background subtraction that initially fits the video to a static background transformed with per-frame homographies. 
This variant of background subtraction is much more robust to camera rotation, but still struggles with camera translation or motion of background elements.

\subsection{Neural Matting}
Since the emergence of deep learning, many efforts have been put into how to improve trimap-based matting with the power of neural networks~\cite{cho2016natural, xu2017deep, lu2019indices, hou2019context, forte2020f, li2020natural, tang2019learning}. Meanwhile, the scope of the question has also expanded from image to video~\cite{sun2021deep, lin2022robust}. 
The training data for video matting requires precise labeling of minute details, such as human hairs, making it difficult to acquire at scale. Therefore, methods such as MODNet~\cite{ke2022modnet} first train on a synthetic dataset with ground-truth labels, and then attempt to deal with the domain shift to real data. 
Our approach instead focuses on optimizing for a single input video without training on large external datasets. 
\cite{lin2022robust} uses a Recurrent Neural Network (RNN) to adaptively leverage both long-term and short-term temporal information, and aggregate information needed for deciding the foreground matte. 
The RNN has implicit information about background appearance in its hidden states.
In contrast, we have explicit color and alpha representations for each component.

\subsection{Layered Video Representations}
Layered representations of video are useful for many editing tasks. Layered Neural Atlases~\cite{kasten2021layered}, Marionette~\cite{smirnov2021marionette} and Deformable Sprites~\cite{Ye_2022_CVPR} decompose videos into multiple layers that deform according to transformations that handle non-rigid motion and self-occlusion. However, these methods are limited to objects whose appearance in video can be mapped to a planar mosaic, so they could not explicitly handle changes in appearance over time, for instance due to lighting.

The work most closely related to ours is Omnimatte~\cite{lu2021omnimatte}, which uses a neural network to decompose a video into a set of color and opacity layers. Omnimatte works by first solving for a background layer that approximates the video as a series of homographies applied to a static image. It then uses a neural network to assign residual video content to a specified number of foreground layers, each initialized with a separate mask. In the case of a single foreground layer, this offers an improved version of background subtraction that is robust to camera rotation, and in the case of multiple foreground layers it tends to group foreground content with whichever foreground mask shares similar motion. Omnimatte is able to associate content like shadows with the foreground objects that cast them, making it effective for re-timing tasks, but it often suffers artifacts when combining content from multiple sources. This is because Omnimatte does not attempt to solve for an independent decomposition of the video. More specifically, reconstruction error in a neural decomposition network is easier to maximize when more layers share their color with the input, as this reduces the reconstruction penalty of possible errors in the alpha channel. Therefore, colors of different layers tend to be correlated. Our re-framing of the matting problem addresses encourages a more independent factorization by specifying that each component should represent a version of the scene where the contents of other components have been removed.

\section{Limits of Classical Matting}
\label{sec:classicalmatting}
In this section we review the classic Bayesian formulation of image matting and the assumptions it makes, which help motivate our re-framing of the problem in Section \ref{sec:FactorMatte}. Much of this section summarizes the original analysis of~\shortcite{chuang2001bayesian}, with some updates to account for video matting and subsequent works.
\subsection{Bayesian Formulation}
\label{sec:basicbayesianmatting}
Matting is traditionally posed as inference of the spatially-varying parameter $\AlphaVal{_p}$ from Equation \ref{eq:classicalmatting} over a provided image or video. As this task is under-constrained, it must be solved with a prior, and we can interpret different approaches as different strategies for choosing and applying that prior. To cover video, we consider a tensor form of Equation \ref{eq:classicalmatting}:
\beq
\Video{}=\AlphaVideo{}\circ\ForegroundVideo + (\mathbf{1}-\AlphaVideo{})\circ\BackgroundVideo
\label{eq:videocompositing}
\eeq
where $\Video{}$ is a given video, and $\AlphaVideo$, $\ForegroundVideo$, and $\BackgroundVideo$ are alpha, foreground, and background tensors with dimensions matching $\Video{}$. 
We can formulate the solution to our matting problem as a \emph{maximum a poteriori} (MAP) estimation of our parameters:
\beq
\argmaxover{\ForegroundVideo{},\BackgroundVideo{},\AlphaVideo{}} \;\; P(\ForegroundVideo{},\BackgroundVideo{},\AlphaVideo{}|\Video{})
\eeq
\noindent Applying Bayes rule, we get:
\beq
\argmaxover{\ForegroundVideo{},\BackgroundVideo{},\AlphaVideo{}} \;\;
\frac{\coverbracelines{P(\Video{}|\ForegroundVideo{},\BackgroundVideo{},\AlphaVideo{})}{Likelihood}{under Eq \ref{eq:classicalmatting}}\coverbrace{P(\ForegroundVideo{},\BackgroundVideo{},\AlphaVideo{})}{Prior}}{P(\Video{})}
\label{eq:jointpriorfull}
\eeq
where the likelihood term $P(\Video{}|\ForegroundVideo{},\BackgroundVideo{},\AlphaVideo{})$ generally measures whether the contents of $\Video{}$, $\ForegroundVideo{}$, $\BackgroundVideo{}$, and $\AlphaVideo$ satisfy Equation \ref{eq:videocompositing}, and $P(\ForegroundVideo{},\BackgroundVideo{},\AlphaVideo{})$ is our joint prior over these values. 

\begin{figure}
    \centering
    \includegraphics[width=0.48\textwidth]{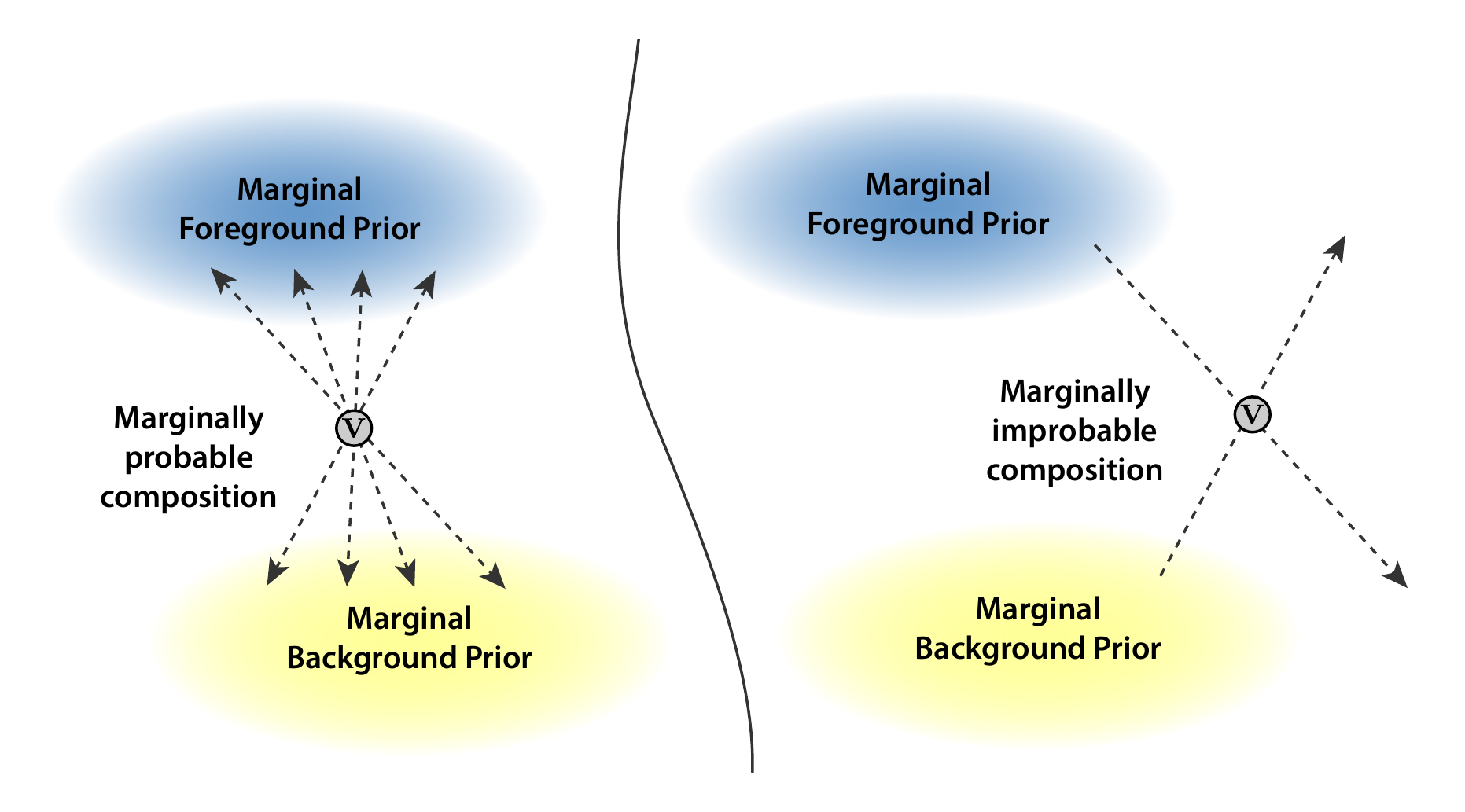}
    \caption{\textbf{Matting likelihood \& marginally improbable interactions}: Consider marginal foreground and background priors over some patch or image space, as used in Equation \ref{eq:independentpriors}. The likelihood in Equation \ref{eq:independentpriors} derives from Equation \ref{eq:videocompositing}, which stipulates that the observation $\Video{}$ should be a linear combination of foreground $\ForegroundVideo$ and background $\BackgroundVideo$. When the observation is a linear combination of marginally probable  $\ForegroundVideo$ and $\BackgroundVideo$ values, likely solutions are also marginally probable (left). However, conditional interactions between the foreground and background can create observations that cannot be explained as a linear combination of marginally probable values (right).}
    \label{fig:likelihood}
\end{figure}

\subsection{Common Assumptions of Classical Matting}
Equation \ref{eq:jointprior} can be simplified by removing $P(V)$, which has no impact on the outcome of our optimization. The classical formulation further assumes a uniform prior on alpha values independent of foreground and background color, and subsequent approaches have either maintained this assumption or implicitly incorporated joint information about the alpha channel into foreground and background priors trained on external data. These observations further simplify Equation \ref{eq:jointpriorfull} to 
\beq
\argmaxover{\ForegroundVideo{},\BackgroundVideo{},\AlphaVideo{}} \;\;
\coverbrace{P(\Video{}|\ForegroundVideo{},\BackgroundVideo{},\AlphaVideo{})}{Likelihood}\coverbrace{P(\ForegroundVideo{},\BackgroundVideo{})}{Prior}
\label{eq:jointprior}
\eeq

Most matting approaches also assume solutions that do not satisfy Equation \ref{eq:videocompositing} have very low likelihood. As the values $\ImageVal{_p}\in\Video{}$ are observed, this effectively restricts the foreground and background values for each pixel, $\{\Foreground{_p}\in\ForegroundVideo{},\Background{_p}\in\BackgroundVideo{}\}$ to pairs that contain $\ImageVal{_p}\in\Video{}$ in their convex span (defined by adherence to the per-pixel Equation \ref{eq:classicalmatting}). Even with this restriction, the matting problem remains under-constrained, yielding many possible solutions (illustrated in Figure \ref{fig:likelihood}, left). This ambiguity can be resolved by maximizing the prior $P(\ForegroundVideo{},\BackgroundVideo{})$ over likely solutions. 
Learning a joint prior $P(\ForegroundVideo{},\BackgroundVideo{})$ directly from $\Video{}$ is usually impractical due to sample complexity. Instead, most matting approaches assume that $\ForegroundVideo{}$ and $\BackgroundVideo{}$ are independent, which allows us to separate the joint distribution over these variables into a product of marginal priors:
\beq
\argmaxover{\ForegroundVideo{},\BackgroundVideo{},\AlphaVideo{}} \;\;
\coverbrace{P(\Video{}|\ForegroundVideo{},\BackgroundVideo{},\AlphaVideo{})}{Likelihood}\coverbracelines{P(\ForegroundVideo{})P(\BackgroundVideo{})}{Marginal}{priors}
\label{eq:independentpriors}
\eeq
\noindent 
The use of marginal priors here makes sense if our end goal is re-composition: Equation \ref{eq:classicalmatting} assumes that we can transform each matted component independently; factoring our prior in this way simply applies the same assumption to the inference of those components. This comes with the added benefit that marginal priors are much easier to learn. In most applications, these marginal priors are derived from either a representative sampling of unambiguous pixel labels (e.g., a trimap or labeled strokes) or assumptions about the motion of different layers (e.g., background subtraction, which assumes the background is static).
Unfortunately, both of these strategies fail to account for visible interactions between the foreground and the background component, which tend to yield pixels with ambiguous, non-binary labels. We can understand these limits as a failure of our marginal priors to capture effects that are only likely when one component is conditioned on its interactions with content from another component.

\subsection{Conditional Interactions}
\begin{figure*}[h]
    \centering
    \captionsetup[subfloat]{labelformat=parens}
    \begin{subfigure}[b]{0.193\textwidth}
         \centering
         \includegraphics[width=1\textwidth]{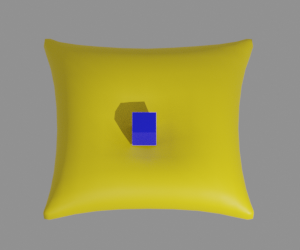}
         \caption{Frame w/ Interaction}
         \label{fig:fig2_static_frame}
     \end{subfigure}
     \begin{subfigure}[b]{0.193\textwidth}
         \centering
         \includegraphics[width=1\textwidth]{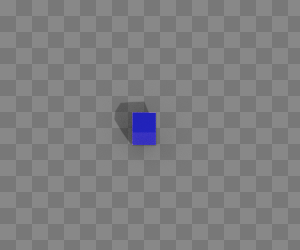}
         \caption{Ideal Foreground}
         \label{fig:fig2_foreground_gt}
     \end{subfigure}
     \begin{subfigure}[b]{0.193\textwidth}
         \centering
         \includegraphics[width=1\textwidth]{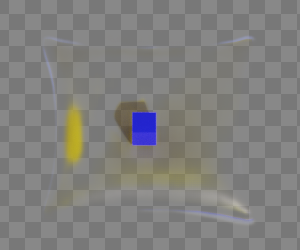}
         \caption{\OmniMatte Foreground}
         \label{fig:fig2_foreground_OM}
     \end{subfigure}
     \begin{subfigure}[b]{0.193\textwidth}
         \centering
         \includegraphics[width=1\textwidth]{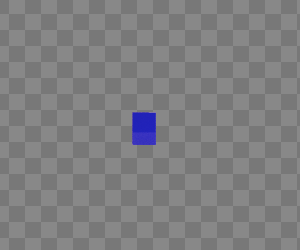}
         \caption{Image Matting Foreground}
         \label{fig:fig2_foreground_IM}
     \end{subfigure}
     \begin{subfigure}[b]{0.193\textwidth}
         \centering
         \includegraphics[width=\textwidth]{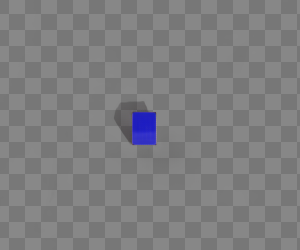}
         \caption{\CausalMatte Foreground}
         \label{fig:fig2_foreground_CM}
     \end{subfigure}
     
     \begin{subfigure}[b]{0.193\textwidth}
         \centering
         \includegraphics[width=1\textwidth]{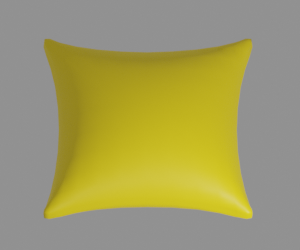}
         \caption{Rest State Background}
         \label{fig:fig2_interaction_frame}
     \end{subfigure}
     \vspace{-0.77\baselineskip}
     \begin{subfigure}[b]{0.193\textwidth}
         \centering
         \includegraphics[width=1\textwidth]{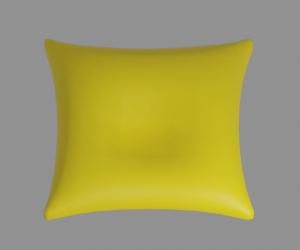}
         \caption{Ideal Background}
         \label{fig:fig2_background_gt}
     \end{subfigure}
     \begin{subfigure}[b]{0.193\textwidth}
         \centering
         \includegraphics[width=1\textwidth]{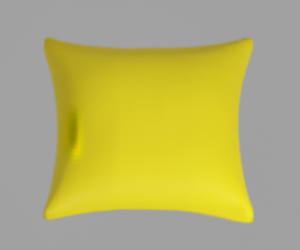}
         \caption{\OmniMatte Background}
         \label{fig:fig2_background_OM}
     \end{subfigure}
     \begin{subfigure}[b]{0.193\textwidth}
         \centering
         \includegraphics[width=1\textwidth]{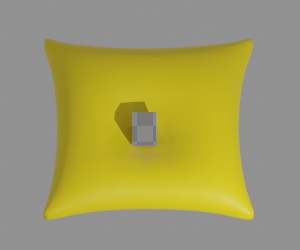}
         \caption{Image Matting Background}
         \label{fig:fig2_background_IM}
     \end{subfigure}
     \begin{subfigure}[b]{0.193\textwidth}
         \centering
         \includegraphics[width=\textwidth]{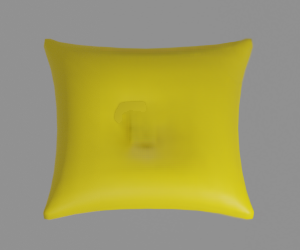}
         \caption{\CausalMatte Background}
         \label{fig:fig2_background_CM}
     \end{subfigure}
      \caption{\textbf{Limits of Classical Matting.} The examples here come from a video of a simulated blue cube bouncing off of a yellow cushion at three different locations. (\subref{fig:fig2_static_frame}) is an input frame when the cube is in contact with and deforming the cushion. The rest of the figure shows a decomposition of this frame into foreground and background content according to different methods. 
      (\subref{fig:fig2_foreground_gt}) and (\subref{fig:fig2_background_gt}) show the ideal factorization we want to achieve. (\subref{fig:fig2_foreground_OM}) and (\subref{fig:fig2_background_OM}) are the results of the video matting method closest to ours \cite{lu2021omnimatte}. (\subref{fig:fig2_foreground_IM}) and (\subref{fig:fig2_background_IM}) show the results of a classical image matting method \cite{chen2013knn}: since such matting methods focus on the foreground matte, the background content behind opaque foreground pixels is undefined.
      (\subref{fig:fig2_foreground_CM}) and (\subref{fig:fig2_background_CM}) show our results. (\subref{fig:fig2_interaction_frame}) shows a static cushion at rest state for better comparison.}
     \label{fig:fig2}
\end{figure*}

The marginal priors in Equation \ref{eq:independentpriors} assume that the appearance of each component is independent. Applied to our foreground and background component, this would mean:
\beq
P(\BackgroundVideo{}|\ForegroundVideo{})=P(\BackgroundVideo{}), \;\; P(\ForegroundVideo{}|\BackgroundVideo{})=P(\ForegroundVideo{})
\label{eq:counterfactual}
\eeq
The statement $P(\BackgroundVideo{}|\ForegroundVideo{})=P(\BackgroundVideo{})$ can be interpreted as saying that the inclusion of foreground content should not change the appearance of background content that remains visible in a composition. Equivalently, our prior on the background should match that of a counterfactual video where the foreground has been removed. However, many common effects violate this rule. For example, the probability of seeing a shadow is much greater if we condition on the presence of some object being included in the scene to cast it. If a foreground object casts a shadow on the background of our video (as in Figure \ref{fig:fig2_static_frame}), this means that the product of marginal priors used in Equation \ref{eq:independentpriors}, will differ from the full joint distribution used in Equation \ref{eq:jointprior}, which can lead the likelihood term in Equation \ref{eq:independentpriors} to favor solutions with low probability under the corresponding priors. This inconsistency becomes even more significant if we consider the situation in Figure \ref{fig:fig2_interaction_frame}, where the foreground cube causes the background cushion to deform. In this case, even if we freeze the scene and remove the foreground cube, the deformed state of the cushion behind it would still be improbable without the explanation of a foreground cube to cause that deformation.

To put these inconsistencies into perspective, consider how they might affect different classical matting strategies. The shadow of our foreground cube in Figure \ref{fig:fig2} derives its color from the background, so appearance-based matting is likely to associate shadow pixels with the background (as in Figure \ref{fig:fig2_background_IM}). On the other hand, the shadow's apparent motion follows that of the cube, so background subtraction and other motion-based techniques are likely to associate shadow pixels with the foreground (as show in Figure \ref{fig:fig2_foreground_OM}). Both of these solutions lead to artifacts in re-compositing, as each leaves color from one component in the contents of another. For re-compositing, a more ideal solution might factor the same shadow pixels into a background representing a counterfactual scene where the cube has been removed (as in Figure \ref{fig:fig2_background_gt}), and a foreground that shows the cube surrounded by colors representing its likely contributions to a new scene. For example, rather than the dark yellow mixture as in Figure \ref{fig:fig2_foreground_OM}, the foreground color contributed by potential shadows of the cube should be pure dark as in Figure \ref{fig:fig2_foreground_gt}. We reformulate the matting problem to favor this type of solution.

\section{The Factor Matting Problem}
\label{sec:FactorMatte}
We propose \emph{factor matting}, an adjusted framing of the matting problem that leads to factorizations of scenes that are better suited for re-compositing. 
The goal of factor matting is to factor a video into different \emph{components}, where each component shows a counterfactual version of the scene with all but its associated contents removed. For symmetry with previous equations, we formulate this as Bayesian inference and describe the case of factoring a foreground component from the rest of the scene, noting that other components, including the background, follow symmetrically.

Building from Equation \ref{eq:jointprior}, we will forego the usual assumption that $\ForegroundVideo{}$ and $\BackgroundVideo{}$ are independent, which was the source of our problems with conditional interactions, and instead note that by the chain rule
\begin{align}
P(\ForegroundVideo{},\BackgroundVideo{})=P(\ForegroundVideo{}|\BackgroundVideo{})P(\BackgroundVideo{})=P(\BackgroundVideo{}|\ForegroundVideo{})P(\ForegroundVideo{})
\label{eq:chainrule}
\end{align}
Applying this to Equation \ref{eq:jointprior}, we can factor our joint prior into the product of marginal and conditional distributions:
\beq
\argmaxover{\ForegroundVideo{},\BackgroundVideo{},\AlphaVideo{}} \;\;
P(\Video{}|\ForegroundVideo{},\BackgroundVideo{},\AlphaVideo{})
\;\;
P(\ForegroundVideo{}|\BackgroundVideo{})
P(\BackgroundVideo{})
\label{eq:singleconditionalprior}
\eeq

\noindent Here we call the conditional term $P(\BackgroundVideo{}|\ForegroundVideo{})$ and its symmetric counterparts $P(\ForegroundVideo{}|\BackgroundVideo{})$ \emph{conditional priors}. 
Equation \ref{eq:singleconditionalprior} is equivalent to Equation \ref{eq:jointprior}, which
shows that we can remove the limiting assumption of independence from classical matting by replacing one of our marginal priors with a conditional one. To make use of this we need a way to learn and interpret our new conditional prior. 

Like its marginal counterpart, $P(\ForegroundVideo{}|\BackgroundVideo{})$ describes a distribution over the appearance of our foreground. Importantly, while the value of $\BackgroundVideo{}$ impacts the distribution that $P(\ForegroundVideo{}|\BackgroundVideo{})$ describes, it is not a parameter of that distribution; so $P(\ForegroundVideo{}|\BackgroundVideo{})$ is the same for any counterfactual version of our scene where $\ForegroundVideo{}$ is found in the same state. We can therefore interpret our conditional prior as analogous to describing its component frozen in time; the impact of other content is preserved in this frozen state, freeing us to replace that content as we see fit.

To see this logic at play in a background, consider the scene in Figure \ref{fig:teaser}, which shows a child (foreground) splashing through a puddle (background). Classical matting struggles with this example because a splash would be improbable under marginal priors for the background---splashes do not usually appear without something there to create them. However, such a splash would be considerably more probable in a distribution describing the puddle conditioned on its observed interaction with the foot. Such a distribution would be equivalent to one describing the splash frozen in time with the foot removed. We can check this interpretation by treating the frozen state as an entirely new scene with background $\BackgroundVideo{}'$. In this case $P(\BackgroundVideo{}')=P(\BackgroundVideo{}|\ForegroundVideo{})$, so $P(\BackgroundVideo{}'|\ForegroundVideo{})=P(\BackgroundVideo{}')$, and we can think of our conditional prior as a marginal prior over a scene where our now-frozen background is really independent of the foreground.

\subsection{Counterfactual Video}
So far we have connected conditional priors to counterfactual versions of a scene and shown that replacing one of the marginal priors in a Bayesian formulation of matting with a conditional one removes the assumption of independence that breaks down for many common scenes. This does not yet suggest a new framing of the problem: Equation \ref{eq:singleconditionalprior} is still classical matting, only with common limiting assumptions removed. Another issue remains when we optimize solutions under this framing. In real scenes, the contribution of a component to video may be sparse in image space; i.e., the ideal alpha channel for that component will have a large number of zeros. This means little or no supervision for the color of that component for much of the scene. In practice this biases optimization toward solutions that make other criteria easier to satisfy, typically resulting in correlated color channels across different components (for examples of this, see the RGB layers produced by \OmniMatte in our supplemental gallery). This is problematic if our ultimate goal is to use factored components for downstream re-compositing applications. We can address this by biasing the color of these unsupervised pixels toward values they are likely to take in other compositions. This bias may be application or scene dependent and could include things like dark values to represent possible shadows a component might cast in other scenes, or colors representing possible color cast, reflections, or illumination.

\subsection{Comparison to Classical Matting}
One way to incorporate bias toward counterfactual video is by re-interpreting our priors as distributions over likely downstream compositions. From here, it can be useful to consider a more symmetric factorization of our prior into marginal and conditional distributions that is equivalent to Equation \ref{eq:singleconditionalprior}:
\beq
\argmaxover{\ForegroundVideo{},\BackgroundVideo{},\AlphaVideo{}} \;\;
\coverbrace{P(\Video{}|\ForegroundVideo{},\BackgroundVideo{},\AlphaVideo{})}{Likelihood}
\;\;[\coverbracelines{P(\ForegroundVideo{})P(\BackgroundVideo{})}{Marginal}{Priors}
\coverbracelines{P(\BackgroundVideo{}|\ForegroundVideo{})P(\ForegroundVideo{}|\BackgroundVideo{})]}{Conditional}{Priors}
^\frac{1}{2}
\label{eq:conditionalpriors}
\eeq
If we take the log likelihood of this form we get:
\beq
\argmaxover{\ForegroundVideo{},\BackgroundVideo{},\AlphaVideo{}} \;\;
\coverbracelines{\LogL{}(\Video{}|\ForegroundVideo{},\BackgroundVideo{},\AlphaVideo{})}{Reconstruction}{Likelihood}
+\frac{
\coverbracelines{
\LogL{}(\ForegroundVideo{})+\LogL(\BackgroundVideo{})
}{Classical Matting}{Priors}
+\coverbracelines{
\LogL{}(\BackgroundVideo{}|\ForegroundVideo{})+\LogL(\ForegroundVideo{}|\BackgroundVideo{})
}{Conditional}{Priors}}{2}
\label{eq:conditionalloglikelihoods}
\eeq
which shows more clearly the relationship to classical matting. 
The relative weights of each term may vary with our confidence in the supplied priors, so in practice we end up with: 

\begin{align}
\argmaxover{\ForegroundVideo{},\BackgroundVideo{},\AlphaVideo{}} \;\;&
w_{r}\LogL{r}+w_{m}\LogL{m}+w_{c}\LogL{c}
\label{eq:factormattelossformulation}
\end{align}
where
\begin{align*}
\LogL{r} &= \LogL{}(\Video{}|\ForegroundVideo{},\BackgroundVideo{},\AlphaVideo{})\\
\LogL{m} &= \LogL{}(\ForegroundVideo{})+\LogL{}(\BackgroundVideo{})\\
\LogL{c} &= \LogL{}(\BackgroundVideo{}|\ForegroundVideo{})+\LogL{}(\ForegroundVideo{}|\BackgroundVideo{})
\end{align*}
and the weights $\{w_r, w_m, w_c\}$ can be adjusted for different applications. From here, it is easy to see that the main difference with classical matting is the inclusion of conditional priors.

\subsection{\InteractionPriors{}}
What makes factor matting tractable is that the conditional distribution $P(\ForegroundVideo{}|\BackgroundVideo{})$ is simpler to approximate than the joint distribution $P(\ForegroundVideo{},\BackgroundVideo{})$. This is largely because our conditional priors have fewer variables than their joint counterparts, which greatly reduces sampling complexity. However, we still need a way to sample the relevant conditional distributions. In Section \ref{sec:FactorMatte} we describe our strategy for doing this in detail. At a high level, we assume that the impact of a cross-component interaction on individual components can be approximated with some combination of image-space transformations (e.g., warping, blurring, brightness adjustments). 
If this is the case, then we should be able to find some distribution of image transformations that can map samples from a marginal prior to an appropriate conditional one. We call this distribution our \emph{\interactionprior{}}. 
If we choose the feature space we use to represent marginal and conditional priors carefully, then our \interactionprior{} can be somewhat generic and applicable to a wide variety of interactions.

\begin{figure*}
\includegraphics[width=1\linewidth]{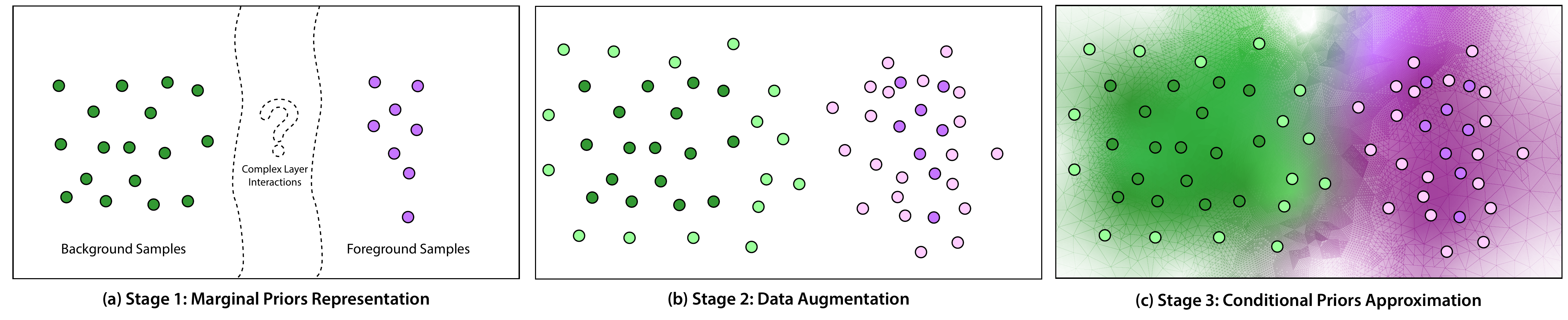}
\caption{\textbf{Data Augmentation for Approximating Conditional Priors} This figure shows an abstract visualization of patch space, where we learn our conditional priors in three stages. In Stage 1 (a), we use traditional matting priors (rough input masks, static background assumption) to obtain unambiguous representative samples of each component. This leaves complex cross-component interactions unlabeled, which are in the center region indicated with a question mark. In Stage 2 (b), we augment these unambiguous samples with transformations designed to span the space of deformations induced by cross-component conditional interactions. Finally, in Stage 3 (c) we adversarially train our networks to model underlying conditional distributions using the augmented samples. 
}
\label{fig:distributions}
\end{figure*}

To train a conditional prior, we first obtain representative appearance samples from the marginal distribution of each component. We then augment those samples with transformations drawn from our \interactionprior{}
Finally, we use these augmented samples to train our conditional priors (see Figure \ref{fig:distributions}).

\begin{figure*}[t]
\centering
\includegraphics[width=\textwidth]{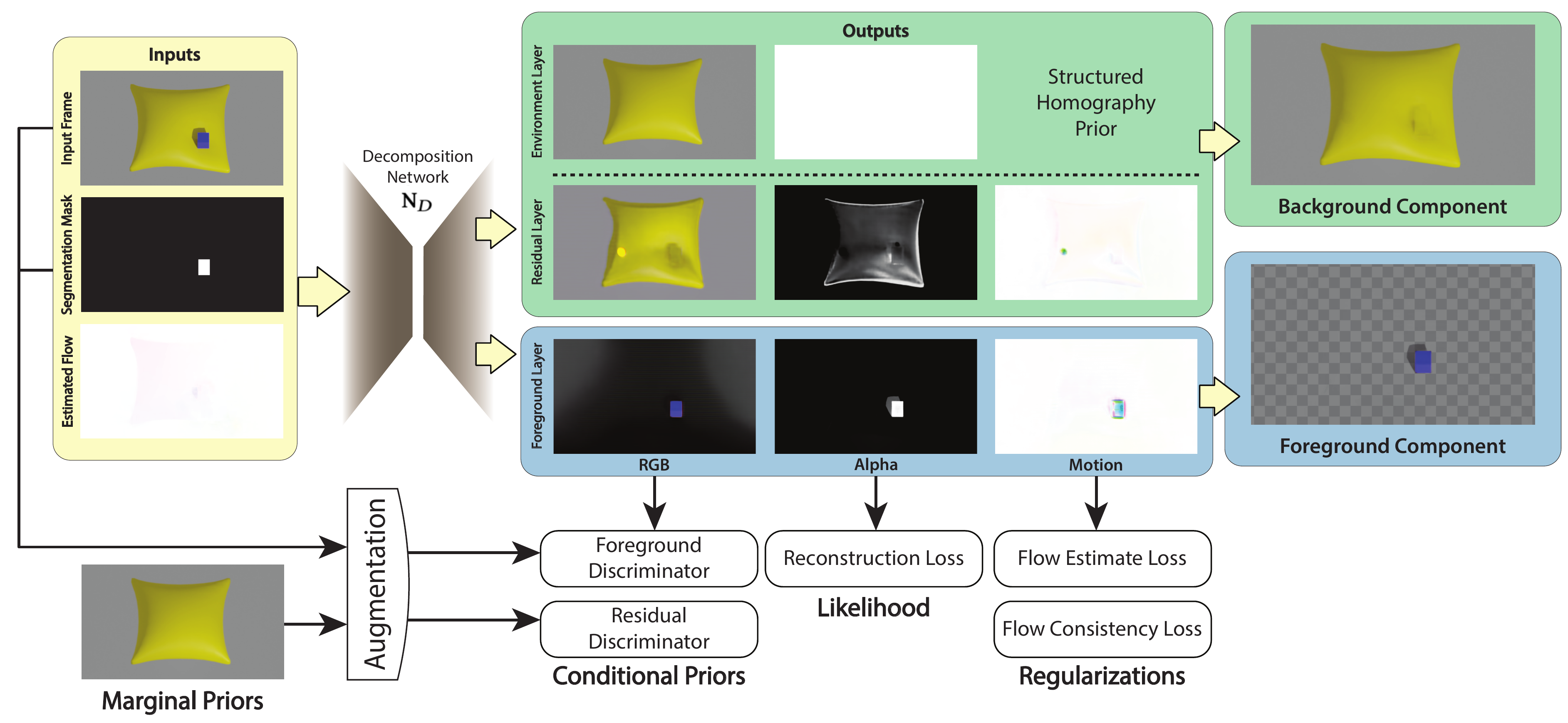}
\caption{\textbf{Method Overview.} Our learning framework takes in a target video and a rough segmentation mask for the object of interests in each frame (left, yellow box), and outputs layers of color and alpha (middle, green and blue boxes). The \FL models the foreground component (right, green box). The \BL and \RL together model the background component (right, blue box). The \BL represents the aspects of the background that are well-described by a homography, while the \RL captures more irregular changes in the background component due, for instance, to interactions with the foreground objects.
Building upon classical Bayesian matting, we use neural networks to represent the probability functions in Equation \ref{eq:independentpriors}, and a gradient-based optimization to optimize for a MAP solution. We use a decomposition network \tdecompN (middle, grey) to produce layers for each component, and supervise it with a reconstruction loss that serves as our likelihood term. Then we define \Structured and \Unstructured priors and instill such knowledge into \tdecompN with adversarially trained discriminators. Finally, we use optical flow information to further regularize the solution.}
\label{fig:method_pipeline}
\end{figure*}

\section{FactorMatte}
\label{sec:methodv1}
We now describe \emph{\OurMethod}, our solution to the factor matting problem. \OurMethod consists of two modules: a decomposition network \tdecompN that factors the input video into one or more layers for each component, and a set of patch-based discriminators $\{\Discr{1},...,\Discr{q}\}$ that represent conditional priors on each component. The reconstruction likelihood $\LogL{r}$ is incorporated as a loss penalizing the difference between each input frame and an ordered composition of the corresponding output layers produced by \tdecompN. For marginal priors, we require them to yield representative samples for each component of a video. Then we augment these representative samples with selected \interactionprior{}s to train a discriminator for each component. The conditional priors $\LogL{c}$ are evaluated by calculating discriminator losses on the outputs of \tdecompN. 

\subsection{Overview}
Our input is a video with $T$ frames and a rough per-frame segmentation mask indicating what content should be associated with the foreground. Our output is a decomposition into color and alpha videos for each component:
\begin{equation*}
    \FactorMatte{}=\{(\Video{\Background{}},\AlphaVideo_{\Background{}}),(\Video{\Foreground{}},\AlphaVideo_{\Foreground{}})\} 
\end{equation*}
For clarity and symmetry with previous methods, we focus on describing the base case of factoring video into one foreground component and one background component. Our method is designed to accommodate more components, but we leave this exploration to future work.

Components are further factored into an ordered set of $N$ layers $\left(\Layer{1},...\Layer{N}\right)$, where each layer is associated with its own priors, and also has its unique position in the composition order, allowing components to interleave. 
Each layer $\Layer{i}=\left(\LayerColor{i},\LayerAlpha{i}\right)$ consists of a color frame at each time,  $\LayerColor{i}=\left(\LayerColor{i}(1),...,\LayerColor{i}(T)\right)$ and corresponding alpha channels  $\LayerAlpha{i}=\left(\LayerAlpha{i}(1),...,\LayerAlpha{i}(T)\right)$. Our default implementation uses one \emph{\FL} for the foreground component, and two layers for the background component: an \emph{\BL} representing static parts of the background, and a \emph{\RL} representing all other, more irregular, aspects of the background. This gives us a total of $N=3$ layers.

Our optimization has three stages, as illustrated in Figure \ref{fig:distributions}. 
In \emph{Stage 1}, we use marginal priors to obtain representative samples for each component. 
In \emph{Stage 2}, we augment these samples with our \interactionprior{}s
to approximate the conditional appearance of each component. Finally in \emph{Stage 3}, we adversarially train discriminators along with the decomposition network \tdecompN to learn conditional distributions over augmented samples, leading to a more independent set of factors derived from the input video.

Sections \ref{subsec:marginalPriors}-\ref{subsec:regularization} describe a default implementation of our method, which we use in our comparisons with previous work. Then in Section \ref{sec:userinputs} we describe additional steps that a user can take to handle scenes with especially complex cross-component interactions. The entire pipeline is illustrated in Figure \ref{fig:method_pipeline}.

\subsection{Marginal Priors}
\label{subsec:marginalPriors}
Our marginal prior for the foreground is produced by masking each frame with the input segmentation, as the appearance in masked regions belongs to the object of interest. We define the background marginal prior as content that can be described by a static image mosaic transformed by a different homography for each frame. To acquire the background marginal prior samples, in Stage 1 we train the decomposition network \tdecompN without discriminators and use the resulting \BL as the base for following augmentations.

\subsection{Conditional Priors}
\label{subsec:conditionalPriors}
The conditional prior for each component is represented with a separate discriminator. These discriminators operate on some feature space, and we must balance the complexity of that feature space against the limited sampling provided by our marginal and \interactionprior{}s. It is illustrative to consider some possible extreme choices: if we simply unrolled the entire image into one long vector, then the dimensionality of our feature space would be enormous, and our sampling of that space would be comparatively minuscule.
Alternatively, if we treated each individual pixel as a distinct sample then our feature space would have only three dimensions and the marginal priors would provide many samples, but the distribution of samples from different components would be likely to overlap. We balance these two failure modes by having our discriminators operate on patch space using an architecture similar to a multi-scale version of PatchGAN~\cite{isola2017image, karnewar2020msg}. 

It is also useful to consider how our choice of feature space relates to our \interactionprior{}s. To be effective, our \interactionprior{}s need to span the projection of deformations caused by cross-component interactions into our feature space. If our feature space were the entire image this could require very scene-specific \interactionprior{}s. For example, to recover the splash in Figure \ref{fig:teaser} our \interactionprior{}s would have to include transformations that span the precise shape of the splash. By contrast, training our discriminators on patches allows us to span these deformations with much more generic transformations (e.g., simple blurring and warping operations). These transformations are applied in Stage 2 of our training, and the details can be found in Section \ref{sec:training}. 

In Stage 3, we train a set of discriminators adversarially along with the decomposition network \tdecompN. We set up one discriminator for the \RL to learn the background marginal priors , and similarly one for the \FL.
For discriminator $\Discr{i}$, its \posex $X^+_i$ are the augmented data is generated in Stage 2, and its negative examples $X^-_i$ are color outputs from \tdecompN. The adversarial loss is:
\begin{equation}
    \Ladv^i = \mathop{\mathbb{E}}\left[\log\left(\Discr{i}(X^+_i)\right)\right]+\mathop{\mathbb{E}}\left[\log\left(1-\Discr{i}(X^-_i)\right)\right]
\end{equation}

\subsection{Likelihood}
The adjusted framing of factor matting puts less importance on Bayesian likelihood than traditional matting. We reflect this in the weight of our reconstruction loss, which is given as:
\begin{equation}
\label{eq1}
    \Lrecon = \frac{1}{T}\sum_{t}||\Video{}(t)-\textbf{Comp}(t)||_{1}
\end{equation}
where \textbf{Comp} is obtained by applying the over blending operator~\cite{porter1984compositing} to the global ordering of layers taken from all components. Our default ordering places the background \BL in the back, followed by the \RL in the middle and the \FL in front. 

\subsection{Regularization}
\label{subsec:regularization}
To further refine the solution, we impose regularization on the alpha and flow outputs. First, we use flow estimation as an auxiliary task for the decomposition network (with corresponding loss \tLest), and use the learned flow to penalize temporal incoherence in color (\tLCwarp) and alpha (\tLAwarp).
Second, we encourage alpha sparsity to prevent repeated content across multiple layers (\tLsparsity). 
Third, to speed up the initial convergence of \FL's alpha channel, we encourage it to match the segmentation mask in the beginning of the training (\tLmask). We turn off this loss once it is below a given threshold. The last two losses are very similar to those used in Omnimatte~\shortcite{lu2021omnimatte}, and we provide details in the Appendix.

\subsubsection{Flow Estimation}
For each layer $\Layer{i}$, the network predicts a flow vector $\LayerFlow{i}(t,t+k)$ from time step $t$ to $t+k$. When $k=1$, it is the most common case of consecutive frames. We use RAFT~\cite{teed2020raft} to estimate the ``ground truth'' flow vector $\LayerGTFlow(t,t+k)$. Current flow estimation algorithms assume that there is just a single mode of flow at each pixel location. However, when complex interactions occur between layers, different flow fields may overlap; for example, a background pixel shadowed by a foreground object might exhibit some evidence of motion correlated with the foreground and some evidence correlated with the background. In such cases, the uni-modal estimate is usually dominated by the layer with the most obvious motion. Therefore, rather than using the estimated ground truth to supervise all layers, we use it to supervise a single layer per pixel. At each pixel location, we find the layer for which the decomposition network assigns the flow vector closest to ground truth, and penalize it according to its deviation from ground truth:
\begin{equation}
    \Lest = \frac{1}{T}\sum_{t}W(t, t+k)\cdot \min_{i}||\LayerFlow{i}(t, t+k)-\LayerGTFlow(t, t+k)||_1
\end{equation}
where the $\min$ operation pools pixel-wise the layer with the least error. $W(t, t+k)$ is a spatial weighting map that reduces the contribution of pixels with inaccurate flow, measured based on standard left-right flow consistency and photometric errors~\shortcite{lu2021omnimatte}.

\subsubsection{Temporal Coherence}
An ideal factorization should produce layers that are temporally coherent. While the decomposition network already satisfies this aim to a degree, we found that adding explicit regularization improved results. In particular, we define losses:
\begin{align}
    \LAwarp &= \frac{1}{T}\frac{1}{N}\sum_{t}\sum_{i}||\LayerAlpha{i}(t)-\textit{warp}\left(\LayerAlpha{i}(t+k), \LayerFlow{i}(t,t+k)\right)||_1\\
    \LCwarp &= \frac{1}{T}\frac{1}{N}\sum_{t}\sum_{i}||\LayerColor{i}(t)-\textit{warp}\left(\LayerColor{i}(t+k), \LayerFlow{i}(t,t+k)\right)||_1 
\label{eq:temporalloss}
\end{align}
The \textit{warp} function warps layer $\Layer{i}(t)$ by predicted flow $\LayerFlow{i}(t, t+k)$. This loss is an extension of one used in Omnimatte~\shortcite{lu2021omnimatte} for non-consecutive frames.

\subsection{Optional Customization}
\label{sec:userinputs}
We find that for complex scenes, results could be improved with very simple additional user input. 
Namely, many videos contain a range of cross-component interactions that take place over time, with some interactions creating more significant deformations than others. In such cases, applying our method first to a subset of frames with less complicated interactions can yield cleaner decompositions of those frames. These clean decompositions can then supply a subsequent optimization, performed over the entire video, with stronger marginal priors and high-confidence reference alpha mattes for regularization. 
To employ this strategy, the user only needs to select a simpler subset of video frames to factor first after Stage 2. 
Then, when training on the entire video in Stage 3, we add a L1 loss penalizing differences from the initial solution on corresponding frames. 

We can further increase the benefit of this strategy by manually cleaning a few frames of alpha produced in this first pass. These improvements can then be propagated to the rest of the video through our regularization terms. 
With the exception of the splash example in Figure \ref{fig:teaser} and the trampoline in Figure \ref{fig:results_tramp}, all of the decompositions shown in this paper were created without this manual intervention step. 
For the splash scene, the alpha is cleaned up on frames preceding the first splash. For the trampoline video, the alpha is cleaned up on frames where the jumper is suspended in air.

When we use high-confidence frames to regularize our optimization over the rest of the video, it helps to include larger values of $k$ in Equation \ref{eq:temporalloss}, as this helps propagate higher-confidence information deeper into parts of the video featuring more complex interactions.

\section{\InteractionPriors{}s}
\label{sec:training}
Our \conditionalmapping{} maps the representative samples provided by our marginal priors to the \posex we supply to each discriminator. 
There are a few generic choices that we use by default for all layers: brightness adjustments, Gaussian blur, and addition of Gaussian noise. These transformations serve the dual purpose of approximating some possible conditional interactions and also providing general data augmentation for improving robustness. In addition to these default transformations, we design some specific ones for the foreground and the background. Each of these transformations is applied at random with a pre-defined probability. The user can customize the algorithms for their own videos and use cases. We emphasize that our contribution here is more on the general framework rather than the specific augmentations used.

\subsection{Foreground Transformations}
\label{obs:training_FL}

\namedparagraph{Shadows}
When one component casts a shadow on another, its contribution to pixels in the other layer is to darken pixels in that other layer. Therefore, to simulate the affect of casting shadows on another layer, we can fill in unsupervised regions with black or dark gray colors.

\namedparagraph{Reflections \& Color Casts}
These effects tend to impart colors from the foreground object onto the background, sometimes darkened by imperfect reflection. To create patches that span this sort of reflection, we can linearly interpolate between copies of the object and patches of pure black.

In summary, in our default implementation the foreground \posex are generated by:
\begin{enumerate}
    \item Randomly choosing an input frame and extracting the object of interest using the provided mask,
    \item Randomly rotating and flipping the extracted region,
    \item Randomly scaling the color of the extracted region,
    \item Filling empty regions uniformly with a random grey value,
    \item Randomly applying the additional generic augmentations described above.
\label{procedure_FL}
\end{enumerate}
This process yields a reasonable approximation to shadows, reflections, and color casts, which are the most common conditional effects of the foreground on other components.

\subsection{Background Transformations}
\begin{figure}
     \centering
     \begin{subfigure}[h]{0.15\textwidth}
         \centering
         \includegraphics[width=\textwidth]{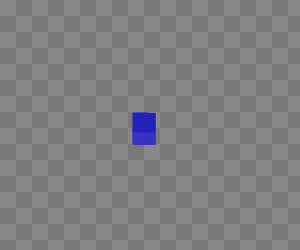}
         \vspace{-1.3\baselineskip}
         \caption{Marginal Prior}
         \label{fig:training_posex_FL_start}
     \end{subfigure}
     \begin{subfigure}[h]{0.15\textwidth}
         \centering
         \includegraphics[width=\textwidth]{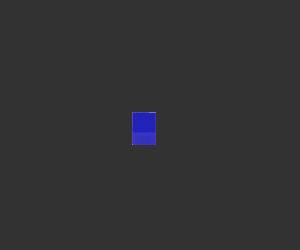}
         \vspace{-1.3\baselineskip}
         \caption{Shadow}
         \label{fig:training_posex_FL_shadow}
     \end{subfigure}
     \begin{subfigure}[h]{0.15\textwidth}
         \centering
         \includegraphics[width=\textwidth]{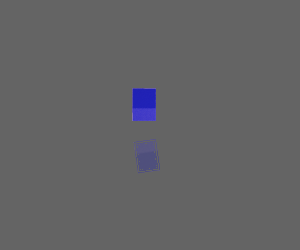}
         \vspace{-1.3\baselineskip}
         \caption{Shadow \& Reflection}
         \label{fig:training_posex_FL_reflec}
     \end{subfigure}
\par\medskip
     \begin{subfigure}[h]{0.15\textwidth}
         \centering
         \includegraphics[width=\textwidth]{figures/fig2/50300_135435crop_static_invis_gt_0080.png}
         \vspace{-1.3\baselineskip}
         \caption{Marginal Prior}
         \label{fig:training_posex_BL_start}
     \end{subfigure}
     \begin{subfigure}[h]{0.15\textwidth}
         \centering
         \includegraphics[width=\textwidth]{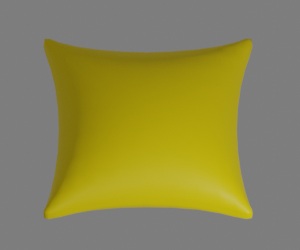}
         \vspace{-1.3\baselineskip}
         \caption{Darkening}
         \label{fig:training_posex_BL_dark}
     \end{subfigure}
     \begin{subfigure}[h]{0.15\textwidth}
         \centering
         \includegraphics[width=\textwidth]{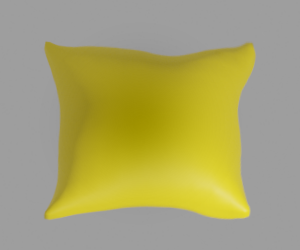}
         \vspace{-1.3\baselineskip}
         \caption{Warping \& Shading}
         \label{fig:training_posex_BL_warp}
     \end{subfigure}

     \caption{\textbf{\Posex for the \FL and \RL.} We show samples from the \posex generated with the default pipeline for the cushion video in Figure \ref{fig:fig2}. For the \FL, we first use the segmentation mask to extract the object of interest. For the remaining empty area, which serves as a representation of the marginal prior (\subref{fig:training_posex_FL_start}). Then to approximate shadow color, (\subref{fig:training_posex_FL_shadow}) fills it with a random grey value. Beyond that, (\subref{fig:training_posex_FL_reflec}) adds a semi-transparent, flipped and rotated copy of the object to simulate reflection. If there is no reflection in the scene, the related augmentation is skipped to further refine the solution space.
     For \RL, after the homography canvas is acquired from Stage 1 as in (\subref{fig:training_posex_BL_start}), (\subref{fig:training_posex_BL_dark}) uniformly darkens it and then adds Gaussian noise. (\subref{fig:training_posex_BL_warp}) applies grid-based warping and adds a Gaussian shade randomly on the cushion. }
     \label{fig:pos_ex}
\end{figure}

Our \BL represents marginally likely parts of the background, forcing conditional background effects into our \RL. We generate positive background samples by applying transformations to the background canvas generated in Stage 1. We use a grid-based warping function that applies a random translation to each vertex in a low-resolution regular grid, upsamples these translations to the resolution of our input video with a Gaussian kernel to generate a smooth displacement field, and applies this displacement field as a warp to a given image. We then randomly darken randomly selected Gaussian regions of the image to approximate bending and self-shading effects, and finally apply the aforementioned generic augmentations. Figure \ref{fig:pos_ex} shows some final outputs from the augmentation pipeline.

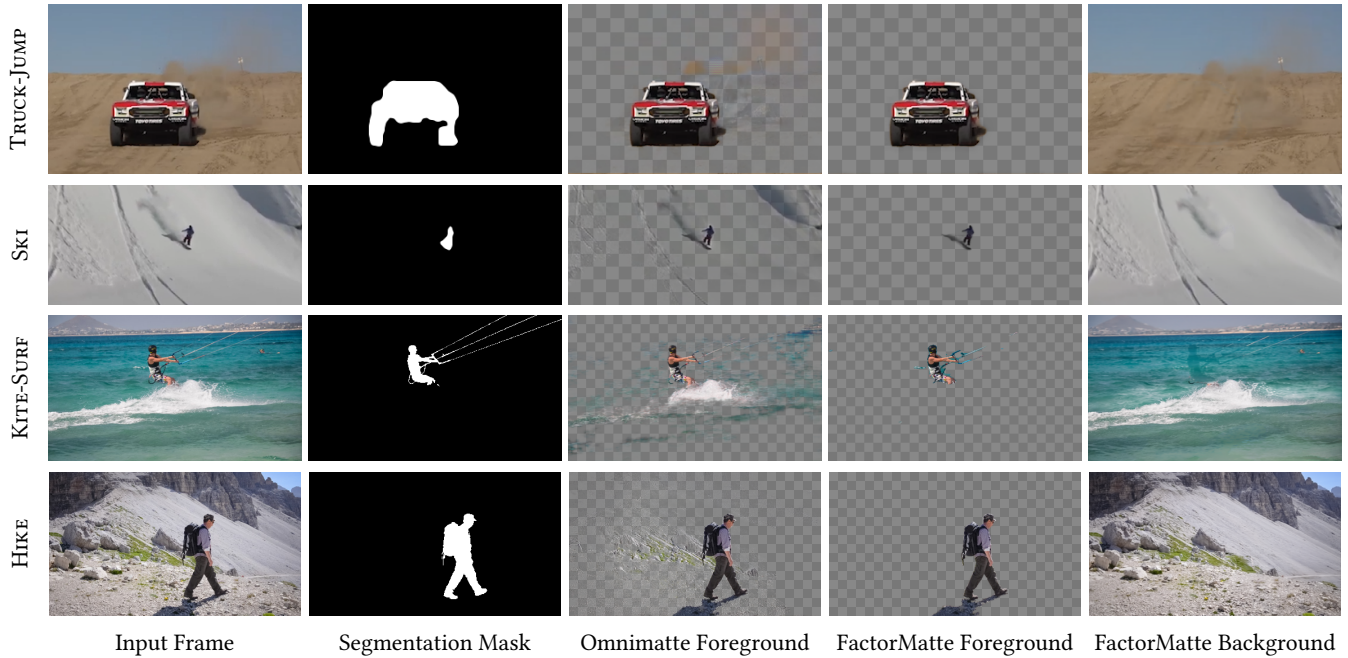
\begin{figure*}
\begin{tikzpicture}
  \node (img01)  {\includegraphics[width=0.188\textwidth]{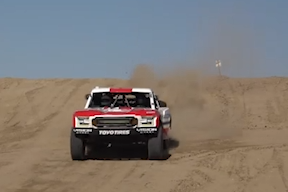}};
  \node[left of=img01, node distance=0.16\textwidth, rotate=90, anchor=center,yshift=-0.045\textwidth,font=\color{black}] {\textsc{Truck-Jump}};
  
  \node[right of=img01,xshift=0.137\textwidth] (img02)  {\includegraphics[width=0.188\textwidth]{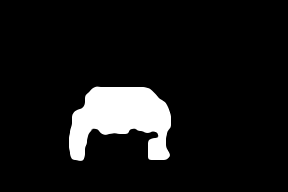}};
  
  \node[right of=img02,xshift=0.137\textwidth] (img03)  {\includegraphics[width=0.188\textwidth]{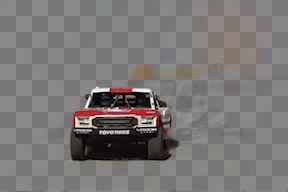}};
  
  \node[right of=img03,xshift=0.137\textwidth] (img04)  {\includegraphics[width=0.188\textwidth]{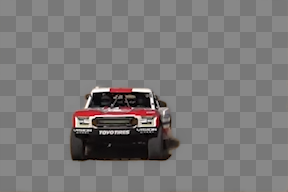}};
  
  \node[right of=img04,xshift=0.137\textwidth] (img05)  {\includegraphics[width=0.188\textwidth]{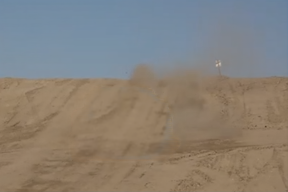}};
  
  \node[below of=img01,yshift=-0.06\textwidth] (img11)  {\includegraphics[width=0.188\textwidth]{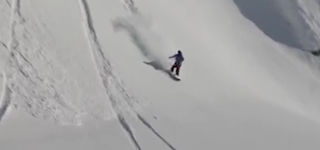}};
  \node[left of=img11, node distance=0.16\textwidth, rotate=90, anchor=center,yshift=-0.045\textwidth,font=\color{black}] {\textsc{Ski}};
  
  \node[right of=img11,xshift=0.137\textwidth] (img12)  {\includegraphics[width=0.188\textwidth]{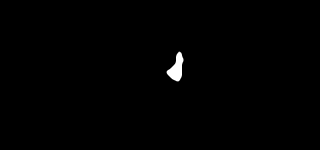}};
  
  \node[right of=img12,xshift=0.137\textwidth] (img13)  {\includegraphics[width=0.188\textwidth]{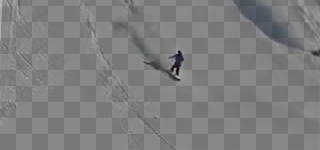}};
  
  \node[right of=img13,xshift=0.137\textwidth] (img14)  {\includegraphics[width=0.188\textwidth]{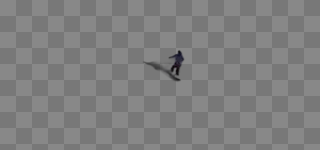}};
  
  \node[right of=img14,xshift=0.137\textwidth] (img15)  {\includegraphics[width=0.188\textwidth]{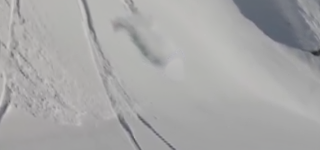}};

  \node[below of=img11,yshift=-0.05\textwidth]  (img21)  {\includegraphics[width=0.188\textwidth]{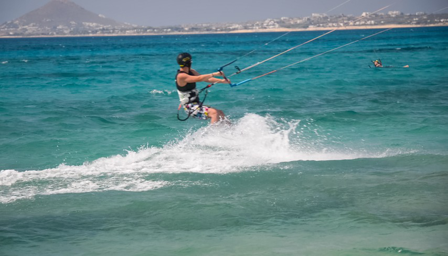}};
  \node[left of=img21, node distance=0.16\textwidth, rotate=90, anchor=center,yshift=-0.045\textwidth,font=\color{black}] {\textsc{Kite-Surf}};
  
  \node[right of=img21,xshift=0.137\textwidth] (img22)  {\includegraphics[width=0.188\textwidth]{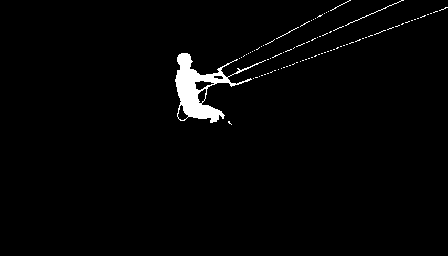}};
  
  \node[right of=img22,xshift=0.137\textwidth] (img23)  {\includegraphics[width=0.188\textwidth]{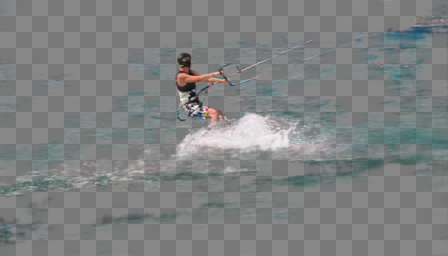}};
  
  \node[right of=img23,xshift=0.137\textwidth] (img24)  {\includegraphics[width=0.188\textwidth]{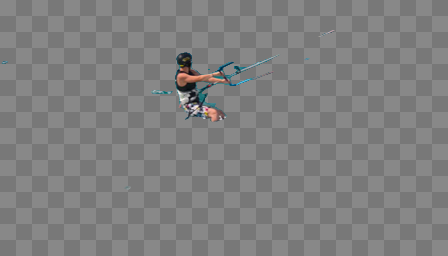}};
  
  \node[right of=img24,xshift=0.137\textwidth] (img25)  {\includegraphics[width=0.188\textwidth]{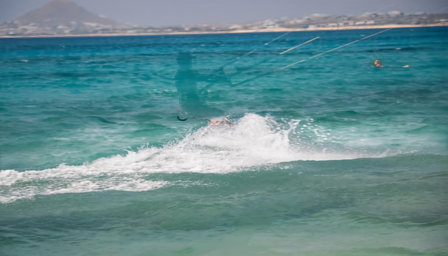}};

  \node[below of=img21,yshift=-0.06\textwidth] (img31)  {\includegraphics[width=0.186\textwidth]{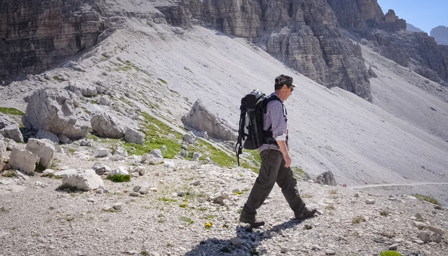}};
  \node[below of=img31, node distance=0.135\textwidth, yshift=0.06\textwidth,font=\color{black}] {Input Frame};
  \node[left of=img31, node distance=0.16\textwidth, rotate=90, anchor=center,yshift=-0.045\textwidth,font=\color{black}] {\textsc{Hike}};
  
  \node[right of=img31,xshift=0.137\textwidth] (img32)  {\includegraphics[width=0.186\textwidth]{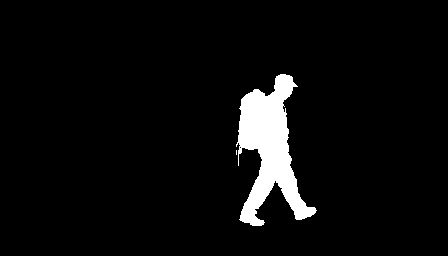}};
  \node[below of=img32, node distance=0.135\textwidth, yshift=0.06\textwidth,font=\color{black}] {Segmentation Mask};
  
  \node[right of=img32,xshift=0.137\textwidth] (img33)  {\includegraphics[width=0.186\textwidth]{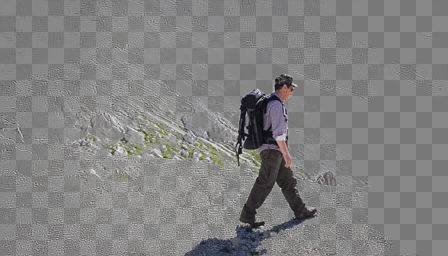}};
  \node[below of=img33, node distance=0.135\textwidth, yshift=0.06\textwidth,font=\color{black}] {\OmniMatte Foreground};
  
  \node[right of=img33,xshift=0.137\textwidth] (img34)  {\includegraphics[width=0.186\textwidth]{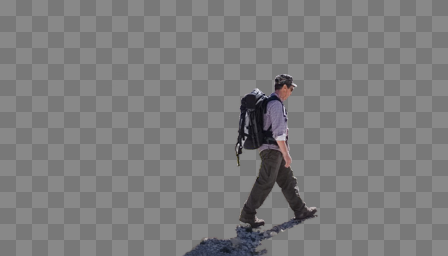}};
  \node[below of=img34, node distance=0.135\textwidth, yshift=0.06\textwidth,font=\color{black}] {\CausalMatte Foreground};
  
  \node[right of=img34,xshift=0.137\textwidth] (img35)  {\includegraphics[width=0.186\textwidth]{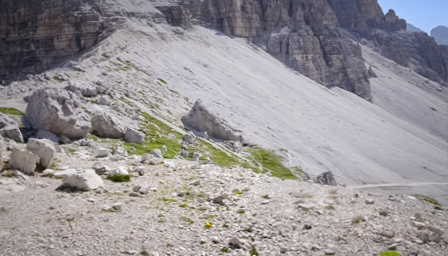}};
  \node[below of=img35, node distance=0.135\textwidth, yshift=0.06\textwidth,font=\color{black}] {\CausalMatte Background};

\end{tikzpicture}
\caption{\textbf{Qualitative comparisons with \OmniMatte.} \CausalMatte separates scenes with and without complex interactions into foreground and background components that are more suitable for re-composition tasks than previous state-of-the-art methods like \OmniMatte. For better visualization of video results please refer to our supplementary materials.}
\label{fig:main}
\end{figure*}
\section{Results}
\label{sec:results}
We test our method on videos with and without complex interactions. Real-world videos featuring interacting elements lack ground truth counterfactual components, so we provide qualitative evaluations on these in Sections \ref{subsec:mainResults}, \ref{subsec:ReComp}, and \ref{subsec:AdditionalEffects}. In Sections \ref{subsec:BS} and \ref{subsec:ablation}, we use datasets and simulated videos that do have ground truth decompositions to provide quantitative comparisons. 
We present results on videos from the DAVIS-2017~\cite{DAVIS2017-1st} and  CDW-2014~\cite{wang2014cdnet} datasets, as well as other common datasets used in specific tasks. 
To include more challenging test cases than those considered in prior work, we also collected clips from in-the-wild videos on YouTube, and recorded additional videos with standard consumer cellphones. Table \ref{table:results_datainfo} describes the test videos we collected. 
For all the methods we evaluate, we use the provided segmentation masks, if any, or else an officially-documented method to automatically generate input masks. 
For the videos we collected, we use a rough binary mask obtained with a quick application of the Rotoscope tool in Adobe After Effects. 
We only manually cleaned up few alpha mattes for Figure \ref{fig:teaser} and \ref{fig:results_tramp} since these examples turned out to be particularly difficult cases for factor matting. 
Our training can be expedited by providing a clean frame featuring the static background without foreground objects or conditional effects. In such cases, Stage 1 is skipped. Acquiring this additional background image is often easy if the users record the input video themselves, so we adopt this practice for videos sourced from cellphone recordings.
\begin{table*}
\begin{center}
    \begin{tabular}{ lllll } 
    \hline
     Location & Name & Source & Description\\ 
     \hline
     Figure \ref{fig:teaser} & \textsc{Puddle Splash} & Youtube & A kid runs into a still puddle, causing water to splash around his foot.\\ 
     Figure \ref{fig:main} & \textsc{Truck-Jump} & YouTube & A truck jumps onto a hill, flinging up dust.\\ 
     Figure \ref{fig:main} & \textsc{Ski} & YouTube & A skier lands on a snowy mountainside, flinging up snow.\\ 
     Figure \ref{fig:inpainting} & \textsc{Poke} & Cellphone Recording & A wooden stick pokes a cushion three times.\\ 
     Figure \ref{fig:color_pop} & \textsc{Flashlight} & Cellphone Recording & A person waves a flashlight, casting a white light beam onto the floor.\\ 
     Figure \ref{fig:fig2}, \ref{fig:method_pipeline} & \textsc{Blue Cube} & Simulation & A blue cube falls onto a yellow cushion and bounces three times.\\ 
     Figure \ref{fig:ablation} & \textsc{Purple Monkey} & Simulation & A purple monkey-head model bounces off a pink textured cushion three times.\\
     \hline
    \end{tabular}
    \caption{\textbf{Description of the videos newly collected in our work.} 
    }
\label{table:results_datainfo}
\end{center}
\end{table*}

\subsection{Complex Scenes}
\label{subsec:mainResults}
In this subsection, we provide a detailed analysis of \OurMethod's results on videos with complex conditional interactions. Our test videos also feature diverse foreground objects that cause a range of lighting effects including hard and soft shadows and reflections. The backgrounds feature a varying opacity, rigidity, texture, shape and motion.

\subsubsection{Videos with Complex Interactions}

Normally the order in which we composite layers, from back to front is: \BL, \RL, and finally \FL. However, as described in Section \ref{sec:methodv1}, the number and order of layers can vary for different scenes to accommodate the interleaving of content from different components. For instance, in Figure \ref{fig:teaser} the transparent water splash is both in front of and behind the child's foot. To represent this, we use two alpha mattes ($\LayerAlpha{\textit{back}}$ and $\LayerAlpha{\textit{front}}$) for the \RL that share the same color channels. Our composition order is then: \BL, $\LayerAlpha{\textit{back}}$ with \RL color, \FL, $\LayerAlpha{\textit{front}}$ with \RL color.
Not only can our model separate out the water surrounding the feet, but also the water right in front of the feet. This task is particularly challenging due to the transparency of the water; the discriminator must learn the flow, color, and texture of the water from the surroundings and apply this knowledge to the region where the water and foot overlap. To our knowledge, no other comparable method can factor this overlapped region into comparably realistic and detailed layers.

The interaction in \textsc{Ski} includes the skier's shadow and the flying snow caused by the skier landing on the mountainside. Since the scene is primarily composed of shades of white or gray, this example is difficult due to significant color overlaps between the foreground shadow and the background mountain trails and flung-up snow. Nonetheless, \OurMethod still produces a clean separation thanks to our use of motion cues.

\begin{figure}[h]
     \centering
          \begin{subfigure}[b]{0.113\textwidth}
         \centering
         \includegraphics[width=1\textwidth]{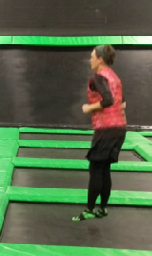}
         \label{fig:results_tramp_input_static}
     \end{subfigure}
     \vspace{-0.77\baselineskip}
     \begin{subfigure}[b]{0.113\textwidth}
         \centering
         \includegraphics[width=1\textwidth]{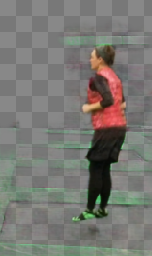}
         \label{fig:results_trampotramp_OM_static}
     \end{subfigure}
     \begin{subfigure}[b]{0.113\textwidth}
         \centering
         \includegraphics[width=1\textwidth]{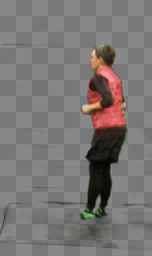}
         \label{fig:results_trampo_CM_FL_static}
     \end{subfigure}
     \begin{subfigure}[b]{0.113\textwidth}
         \centering
         \includegraphics[width=1\textwidth]{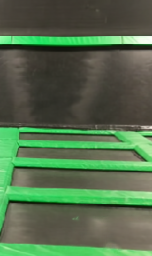}
         \label{fig:results_trampo_CM_RL_static}
     \end{subfigure}

     \begin{subfigure}[b]{0.113\textwidth}
         \centering
         \includegraphics[width=1\textwidth]{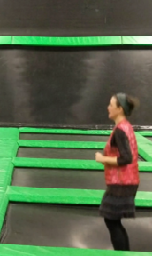}
         \caption{Input Frame \newline}
         \label{fig:results_tramp_input}
     \end{subfigure}
     \begin{subfigure}[b]{0.113\textwidth}
         \centering
         \includegraphics[width=1\textwidth]{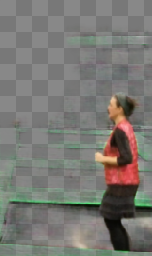}
         \caption{\OmniMatte \protect\\ Foreground}
         \label{fig:results_trampotramp_OM}
     \end{subfigure}
     \begin{subfigure}[b]{0.113\textwidth}
         \centering
         \includegraphics[width=1\textwidth]{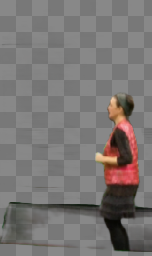}
         \caption{\CausalMatte \protect\\ Foreground}
         \label{fig:results_trampo_CM_FL}
     \end{subfigure}
     \begin{subfigure}[b]{0.113\textwidth}
         \centering
         \includegraphics[width=1\textwidth]{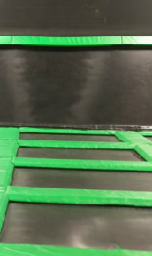}
         \caption{\CausalMatte \protect\\ Background}
         \label{fig:results_trampo_CM_RL}
     \end{subfigure}
     \caption{\textbf{Trampoline (cropped).} The full version of this video is from~\cite{lu2020layered} and~\cite{lu2021omnimatte}, and can be found in Figure \ref{fig:retiming}. The person jumps onto the trampoline and bounces back into the air three times.
     The first row shows a frame where the person is in the air, and the second row shows a frame with the trampoline deformed due to physical interaction. Notice that our foreground is less contaminated by background elements, and our background features more of the trampoline's deformation.}
     \label{fig:results_tramp}
\end{figure}

Conditional interactions are relatively rare among examples used in previous matting works, presumably because most methods were not designed to address such interactions. One partial exception is the work of \OmniMatte ~\shortcite{lu2021omnimatte} and its predecessor~\cite{lu2020layered}, which include examples with shadows, reflections, and some limited deformations. Figure \ref{fig:results_tramp} shows a crop from the trampoline video, one of the most difficult cases for factor matting appearing in their work. 
If we consider the quality of the re-compositing results, both \OmniMatte and \OurMethod produce errors around the edges of the trampoline, yet our method produces a overall cleaner matte that can be used to create a counterfactual video with the jumper removed.

The \textsc{Kite-Surf} video, shown in Figure \ref{fig:main}, depicts a foreground surfer and features significant foreground-background interaction due to the action of surfing. Waves appear in all frames as conditional effects, and the regions of them that can be described by a homography is captured in the \BL during Stage 1. Then the \RL \posex generated from those \BL images also contain some wave colors and textures, which saves us from manually simulating the waves from scratch.

\subsubsection{Background Parallax}
Conditional effects stemming from interactions with the foreground is just one reason that even advanced variations on background subtraction can fail. Other causes include spatially-varying parallax caused by camera translation or changes in scene lighting, which are common phenomena in real videos. One benefit of our conditional priors is that they help to factor these changes into foreground and background effects. For example, the \textsc{Hike} video in Figure \ref{fig:main} was included as a failure case in the \OmniMatte paper~\shortcite{lu2021omnimatte}. This video shows a person walking in front of a mountain, where camera translation and a significant range of scene depths lead to spatially-varying parallax that cannot be approximated with a simple homography. This causes \OmniMatte to place much of the mountain's visual changes in the foreground. By contrast, our method correctly assigns the mountain's visual changes to the \RL and gives a clean foreground containing only the man and his shadow.

\subsection{Background Subtraction}
\label{subsec:BS}
\begin{figure}[t]
     \centering
     \begin{subfigure}[b]{0.15\textwidth}
         \centering
         \includegraphics[width=1\textwidth]{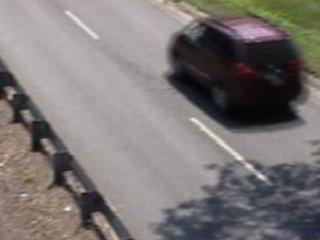}
         \vspace{-1.3\baselineskip}
         \caption{Input Frame}
         \label{fig:BS_traffic_input}
     \end{subfigure}
     \begin{subfigure}[b]{0.15\textwidth}
         \centering
         \includegraphics[width=1\textwidth]{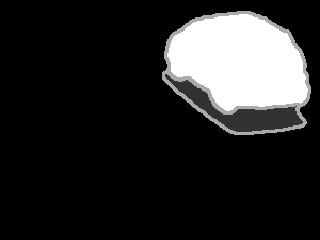}
         \vspace{-1.3\baselineskip}
         \caption{Ground Truth}
         \label{fig:BS_traffic_gt}
     \end{subfigure}
     \begin{subfigure}[b]{0.15\textwidth}
         \centering
         \includegraphics[width=1\textwidth]{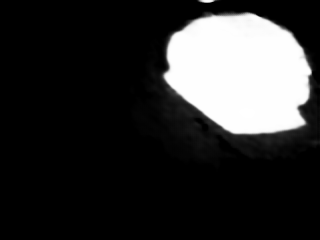}
         \vspace{-1.3\baselineskip}
         \caption{\CausalMatte}
     \end{subfigure}
\par\medskip
     \begin{subfigure}[b]{0.15\textwidth}
         \centering
         \includegraphics[width=1\textwidth]{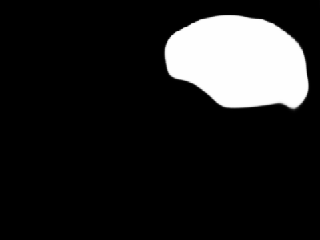}
         \vspace{-1.3\baselineskip}
         \caption{FgSegNet-v2}
     \end{subfigure}
     \begin{subfigure}[b]{0.15\textwidth}
         \centering
         \includegraphics[width=1\textwidth]{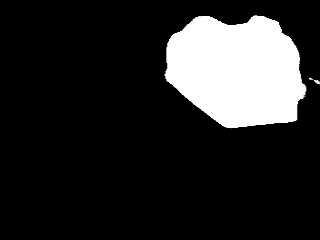}
         \vspace{-1.3\baselineskip}
         \caption{BSUV-Net}
     \end{subfigure}
     \begin{subfigure}[b]{0.15\textwidth}
         \centering
         \includegraphics[width=1\textwidth]{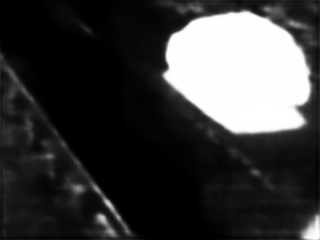}
         \vspace{-1.3\baselineskip}
         \caption{\OmniMatte}
         \label{fig:BS_traffic_OM}
     \end{subfigure}
\par\medskip     
     \begin{subfigure}[b]{0.15\textwidth}
         \centering
         \includegraphics[width=1\textwidth]{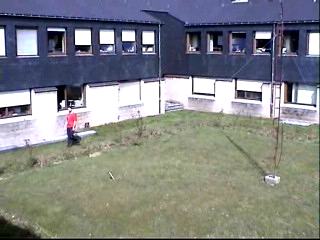}
         \vspace{-1.3\baselineskip}
         \caption{Input Frame}
         \label{fig:BS_zoom_input}
     \end{subfigure}
     \begin{subfigure}[b]{0.15\textwidth}
         \centering
         \includegraphics[width=1\textwidth]{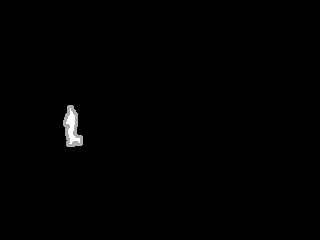}
         \vspace{-1.3\baselineskip}
         \caption{Ground Truth}
         \label{fig:BS_zoom_gt}
     \end{subfigure}
     \begin{subfigure}[b]{0.15\textwidth}
         \centering
         \includegraphics[width=1\textwidth]{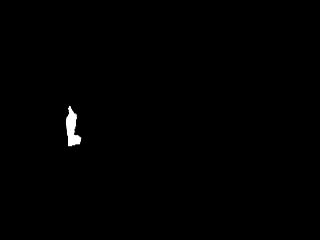}
         \vspace{-1.3\baselineskip}
         \caption{\CausalMatte}
     \end{subfigure}
\par\medskip
     \begin{subfigure}[b]{0.15\textwidth}
         \centering
         \includegraphics[width=1\textwidth]{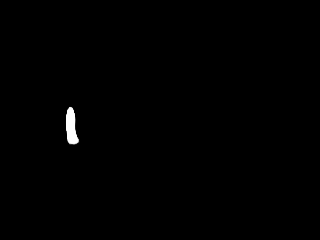}
         \vspace{-1.3\baselineskip}
         \caption{FgSegNet-v2}
     \end{subfigure}
     \begin{subfigure}[b]{0.15\textwidth}
         \centering
         \includegraphics[width=1\textwidth]{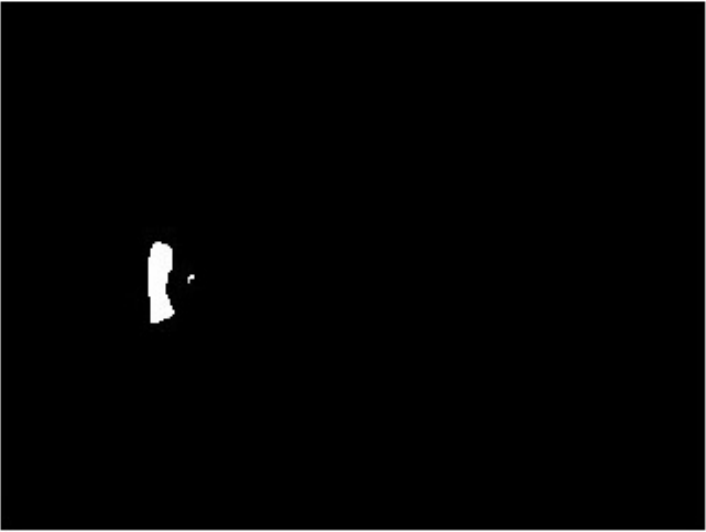}
         \vspace{-1.3\baselineskip}
         \caption{BSUV-Net}
     \end{subfigure}
     \begin{subfigure}[b]{0.15\textwidth}
         \centering
         \includegraphics[width=1\textwidth]{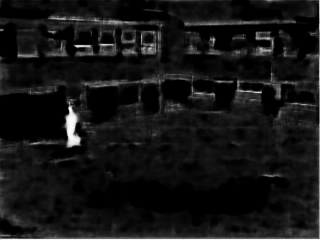}
         \vspace{-1.3\baselineskip}
         \caption{\OmniMatte}
         \label{fig:BS_zoom_OM}
     \end{subfigure}
     
     \caption{\textbf{Background Subtraction.} Parts (\subref{fig:BS_traffic_input}-\subref{fig:BS_traffic_OM}) are from the \textsc{Traffic} video in the \textsc{CameraJitter} category, and parts (\subref{fig:BS_zoom_input}-\subref{fig:BS_zoom_OM}) are from the \textsc{ZoomInZoomOut} video in the \textsc{PTZ} category of CDW-2014. In the Ground Truth (Figure (\subref{fig:BS_traffic_gt}) and (\subref{fig:BS_zoom_gt})), white pixels indicate moving objects, dark grey pixels indicate hard shadows, and light gray pixels indicate unknown motion, typically at boundaries. We set pixels of moving objects and hard shadows as positive labels, and the rest as negative. The results of \CausalMatte and \OmniMatte are the \FL alpha.}
     \label{fig:results_BS}
\end{figure}
We evaluate \CausalMatte on the task of background subtraction with CDW-2014~\shortcite{wang2014cdnet}, a change detection dataset that has pixel-wise labeling of objects, hard shadows, and motions. We focus on clips with shadows and reflections that should be associated with foreground objects, and 
identify 10 such clips from the 11 video categories. The clips feature significant camera jitter, as well as changes in zoom and exposure. As human-annotated labels are provided with this dataset, we quantitatively compare with \OmniMatte and two top-performing background subtraction methods, FgSegNet-v2~\cite{lim2020learning} and BSUV-Net~\cite{tezcan2020bsuv}. 
For each video we train \OmniMatte for 2,000 epochs as indicated in~\cite{lu2021omnimatte}.
We binarize the alpha prediction of all methods by thresholding at 0.8, and report precision, recall, F-score and AUC values in Table \ref{table:results_BS}. Artifacts appearing in the Omnimatte foreground are elements that cannot be expressed by their background, so in order to minimize the reconstruction loss, at times some background elements start appearing in the foreground. Since our \RL is able to capture those irregular background content, our method outperforms \OmniMatte, and is on par with FgSegNet-v2 in terms of recall. We hypothesize our precision is lower because the manually labeled ground truth is relatively conservative: the edges of moving objects are usually labeled as ``unknown’’ as shown in Figures \ref{fig:BS_traffic_gt} and \ref{fig:BS_zoom_gt}. AUC values indicate that \CausalMatte and FgSegNet-v2 are more robust in their predictions against different alpha binarization thresholds, whereas the performance of \OmniMatte significantly deteriorates when the threshold becomes lower, given its prediction contains many small noises.
\begin{table}
\begin{center}
    \begin{tabular}{ lllll } 
    \hline
     Method & Precision & Recall & F-Score & AUC\\ 
     \hline
     \CausalMatte  & 0.767 & 0.773 & 0.770 & 0.934\\ 
     \OmniMatte   & 0.771 & 0.740 & 0.754 & 0.922\\ 
     FgSegNet-v2 & 0.896 & 0.778 & 0.830 & 0.932\\ 
     \hline
    \end{tabular}
    \caption{\textbf{Evaluation on background subtraction.} Our method is comparable to FgSegNet-v2 in terms of recall and AUC.}
\label{table:results_BS}
\end{center}
\end{table}

\subsection{Standard Video Matting}
\begin{figure}
    \centering
    \captionsetup[subfloat]{labelformat=empty}
     \begin{subfigure}[b]{0.15\textwidth}
         \centering
         \includegraphics[width=1\textwidth]{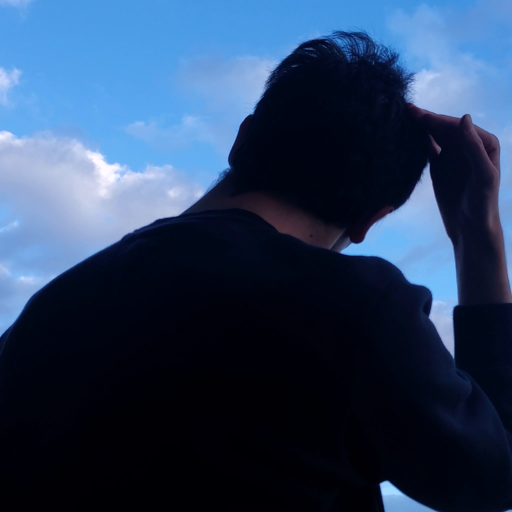}
         \vspace{-1.3\baselineskip}
         \caption{Input Frame}
         \label{fig:VM_gt}
     \end{subfigure}
     \begin{subfigure}[b]{0.15\textwidth}
         \centering
         \includegraphics[width=1\textwidth]{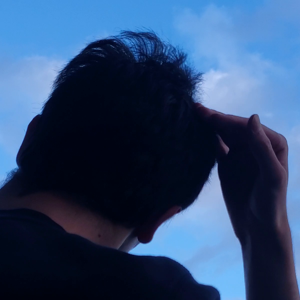}
         \vspace{-1.3\baselineskip}
         \caption{Enlarged View}
         \label{fig:VM_gt_crop}
     \end{subfigure}
     \begin{subfigure}[b]{0.15\textwidth}
         \centering
         \includegraphics[width=\textwidth]{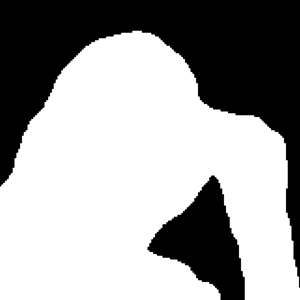}
         \vspace{-1.3\baselineskip}
          \caption{Input Mask}
         \label{fig:VM_mask}
     \end{subfigure}
\par\medskip
     \begin{subfigure}[b]{0.15\textwidth}
         \centering
         \includegraphics[width=1\textwidth]{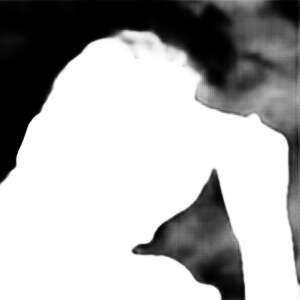}
         \vspace{-1.3\baselineskip}
         \caption{\OmniMatte $\alpha$}
         \label{fig:VM_CM_alpha}
     \end{subfigure}
     \begin{subfigure}[b]{0.15\textwidth}
         \centering
         \includegraphics[width=1\textwidth]{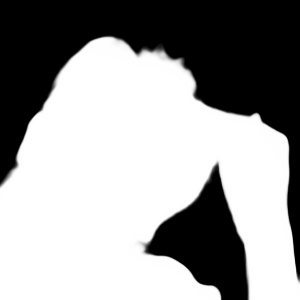}
         \vspace{-1.3\baselineskip}
         \caption{BGM $\alpha$}
         \label{fig:VM_CM_RGBA}
     \end{subfigure}
     \begin{subfigure}[b]{0.15\textwidth}
         \centering
         \includegraphics[width=\textwidth]{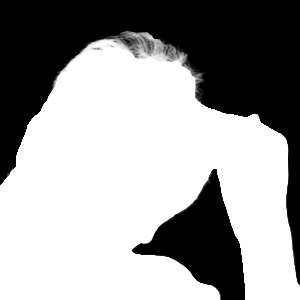}
         \vspace{-1.3\baselineskip}
         \caption{\CausalMatte $\alpha$}
         \label{fig:VM_OM_alpha}
     \end{subfigure}
\par\medskip
     \begin{subfigure}[b]{0.15\textwidth}
         \centering
         \includegraphics[width=1\textwidth]{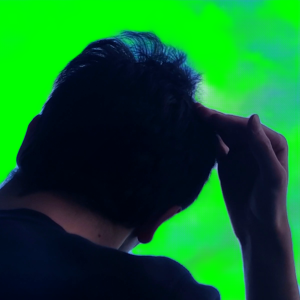}
         \vspace{-1.3\baselineskip}
         \caption{\OmniMatte Matte}
         \label{fig:VM_OM_RGBA}
     \end{subfigure}
          \begin{subfigure}[b]{0.15\textwidth}
         \centering
         \includegraphics[width=1\textwidth]{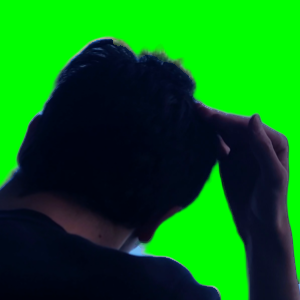}
         \vspace{-1.3\baselineskip}
         \caption{BGM Matte}
         \label{fig:VM_BGM_alpha}
     \end{subfigure}
     \begin{subfigure}[b]{0.15\textwidth}
         \centering
         \includegraphics[width=\textwidth]{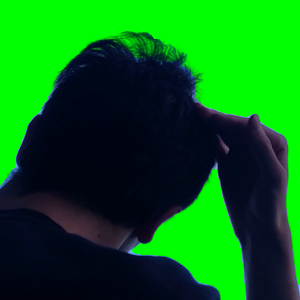}
         \vspace{-1.3\baselineskip}
         \caption{\CausalMatte Matte}
         \label{fig:VM_BGM_RGBA}
     \end{subfigure}
\vspace{-0.77\baselineskip}
\par\smallskip
     
     \hrulefill\par  \smallskip
     \begin{subfigure}[b]{0.15\textwidth}
         \centering
         \includegraphics[width=1\textwidth]{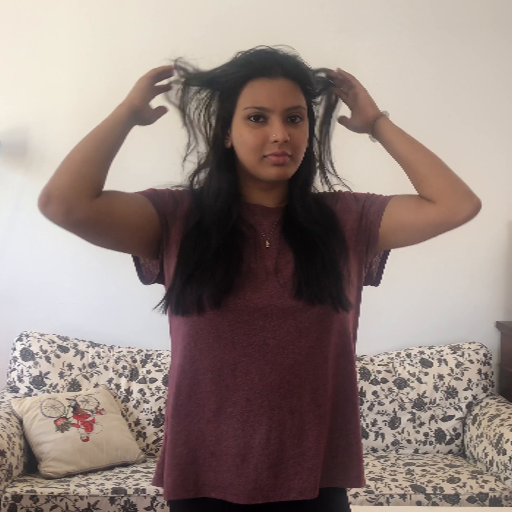}
         \vspace{-1.3\baselineskip}
         \caption{Input Frame}
         \label{fig:VM_0538_gt}
     \end{subfigure}
     \begin{subfigure}[b]{0.15\textwidth}
         \centering
         \includegraphics[width=1\textwidth]{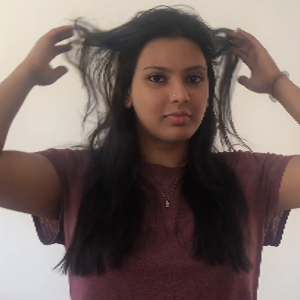}
         \vspace{-1.3\baselineskip}
         \caption{Enlarged View}
         \label{fig:VM_0538_gt_crop}
     \end{subfigure}
     \begin{subfigure}[b]{0.15\textwidth}
         \centering
         \includegraphics[width=\textwidth]{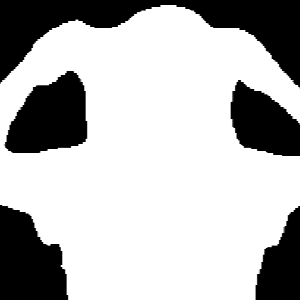}
         \vspace{-1.3\baselineskip}
         \caption{Input Mask}
         \label{fig:VM_0538_mask}
     \end{subfigure}
\par\medskip
     \begin{subfigure}[b]{0.15\textwidth}
         \centering
         \includegraphics[width=1\textwidth]{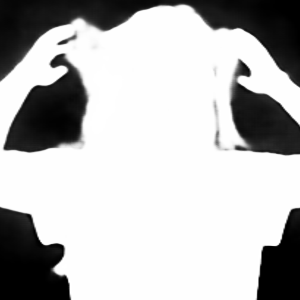}
         \vspace{-1.3\baselineskip}
         \caption{\OmniMatte $\alpha$}
         \label{fig:VM_0538_CM_alpha}
     \end{subfigure}
     \begin{subfigure}[b]{0.15\textwidth}
         \centering
         \includegraphics[width=1\textwidth]{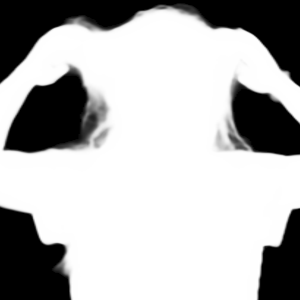}
         \vspace{-1.3\baselineskip}
         \caption{BGM $\alpha$}
         \label{fig:VM_0538_CM_RGBA}
     \end{subfigure}
     \begin{subfigure}[b]{0.15\textwidth}
         \centering
         \includegraphics[width=\textwidth]{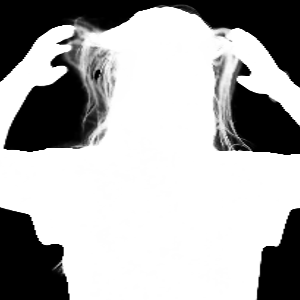}
         \vspace{-1.3\baselineskip}
         \caption{\CausalMatte $\alpha$}
         \label{fig:VM_0538_OM_alpha}
     \end{subfigure}
\par\medskip
     \begin{subfigure}[b]{0.15\textwidth}
         \centering
         \includegraphics[width=1\textwidth]{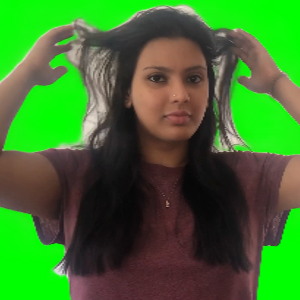}
         \vspace{-1.3\baselineskip}
         \caption{\OmniMatte Matte}
         \label{fig:VM_0538_OM_RGBA}
     \end{subfigure}
     \begin{subfigure}[b]{0.15\textwidth}
         \centering
         \includegraphics[width=1\textwidth]{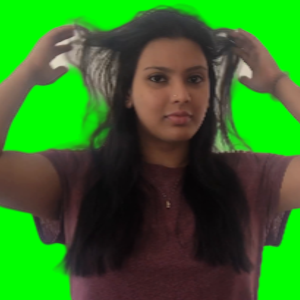}
         \vspace{-1.3\baselineskip}
         \caption{BGM Matte}
         \label{fig:VM_0538_BGM_alpha}
     \end{subfigure}
     \begin{subfigure}[b]{0.15\textwidth}
         \centering
         \includegraphics[width=\textwidth]{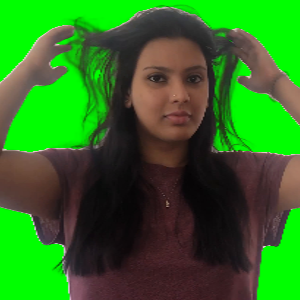}
         \vspace{-1.3\baselineskip}
         \caption{\CausalMatte Matte}
         \label{fig:VM_0538_BGM_RGBA}
     \end{subfigure}
     \caption{\textbf{Video Matting.} Every image except the input frame is an enlarged crop from the raw output. Matte images are the RGBA outputs composited with a green background.}
     \label{fig:results_VM}
     
\end{figure}

While \CausalMatte is designed to address videos featuring complex cross-component interactions, we find that it also excels on scenes without such interactions. 
For example, we evaluate on the published test set from~\cite{lin2021real}, which includes real videos captured by consumer phones fixed on a tripod. Figure \ref{fig:results_VM} compares our results to those of \OmniMatte and Background Matting (BGM)~\cite{sengupta2020background}. Due to memory constraints, BGM fixes the input resolution to 512$\times$512. We stick to the same resolution, and also due to memory limits, we are unable to compare with other methods focusing on HD inputs. In addition to the input video, BGM requires segmentation masks and a static background image. We follow BGM's original practice to use Deeplabv3+ \cite{chen2018encoder} for mask extraction, and use the same set of inputs without extra ones. The static background image is leveraged to generate higher-quality \posex for \RL. For generating \FL \posex, we erode the segmentation masks to avoid background colors. As can be seen in Figure~\ref{fig:results_VM}, \CausalMatte recovers the most accurate hair details among all three methods.

\subsection{After Effects Plugin \& Re-Compositing}
\label{subsec:ReComp}
\begin{figure}[t]
     \centering
     \begin{subfigure}[b]{0.23\textwidth}
         \centering
         \includegraphics[width=1\textwidth]{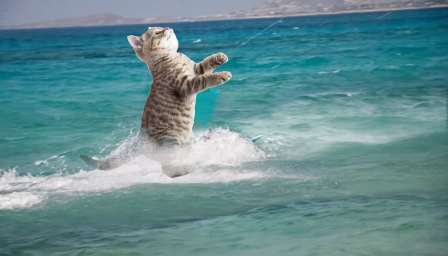}
         \vspace{-1.3\baselineskip}
         \caption{Kite-Surf}
         \label{fig:recomp_surf_CM}
     \end{subfigure}
     \begin{subfigure}[b]{0.23\textwidth}
         \centering
         \includegraphics[width=1\textwidth]{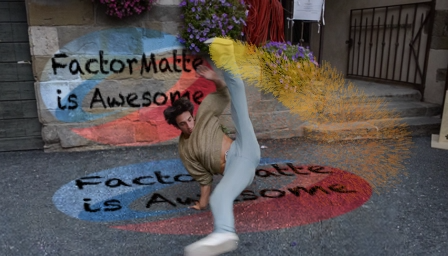}
         \vspace{-1.3\baselineskip}
         \caption{Break-Dance}
         \label{fig:recomp_breakdance_CM}
     \end{subfigure}
\par\medskip
     \begin{subfigure}[b]{0.23\textwidth}
         \centering
         \includegraphics[width=1\textwidth]{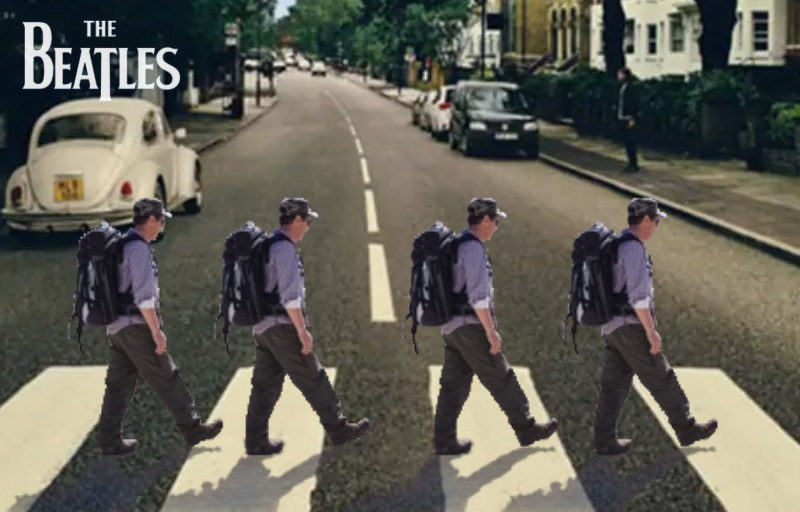}
         \vspace{-1.3\baselineskip}
         \caption{Hikers on Abbey Road}
         \label{fig:recomp_hiker_AR}
     \end{subfigure}
     \begin{subfigure}[b]{0.23\textwidth}
         \centering
         \includegraphics[width=1\textwidth]{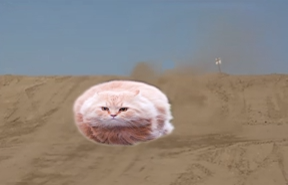}
         \vspace{-1.3\baselineskip}
         \caption{Cat-Jump}
         \label{fig:recomp_hiker_Moon}
     \end{subfigure}
     \caption{\textbf{Re-Composition tasks}. In (\subref{fig:recomp_surf_CM}) and (\subref{fig:recomp_hiker_Moon}) we replace the foreground content while preserving the background conditional effects. In (\subref{fig:recomp_breakdance_CM}) we add additional hand-drawn layers between foreground and background layers. In (\subref{fig:recomp_hiker_AR}) we replace the background. Please see the supplemental materials for full results.}
     \label{fig:recompp}
\end{figure}

\CausalMatte was explicitly designed with re-compositing applications in mind. To facilitate these, we developed a plug-in for Adobe After Effects that loads the output of \CausalMatte into a hierarchy of compositions for editing. These compositions include the components and layers of each scene, as well as the result of binary operators between layers and input masks. The plugin offers several controls and visualization tools, including template compositions with empty layers that users can drag replacement foreground or background objects onto for fast re-composition. Figure \ref{fig:recompp} shows some of the previous examples re-composited with virtual content using our plug-in. We will release this plug-in along with our data and code upon publication of this paper.

\subsection{Additional Video Editing Effects}
\label{subsec:AdditionalEffects}
\begin{figure}[t]
     \centering
     \begin{subfigure}[b]{0.113\textwidth}
         \centering
         \includegraphics[width=1\textwidth]{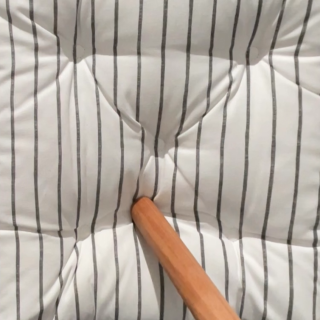}
         \vspace{-1.3\baselineskip}
         \caption{Input Frame}
         \label{fig:inpainting_gt}
     \end{subfigure}
     \begin{subfigure}[b]{0.113\textwidth}
         \centering
         \includegraphics[width=1\textwidth]{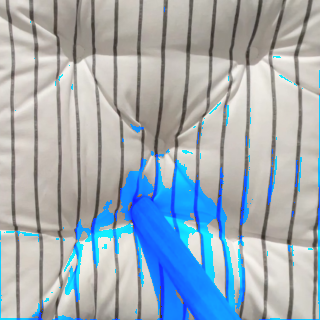}
         \vspace{-1.3\baselineskip}
         \caption{\OmniMatte}
         \label{fig:inpainting_OM_input}
     \end{subfigure}
     \begin{subfigure}[b]{0.113\textwidth}
         \centering
         \includegraphics[width=1\textwidth]{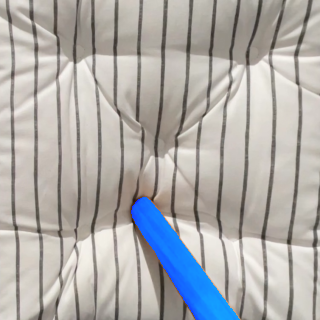}
         \vspace{-1.3\baselineskip}
         \caption{Segmentation}
         \label{fig:inpainting_seg_input}
     \end{subfigure}
     \begin{subfigure}[b]{0.113\textwidth}
         \centering
         \includegraphics[width=1\textwidth]{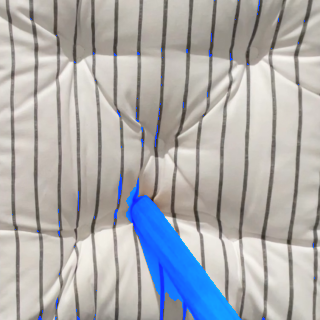}
         \vspace{-1.3\baselineskip}
         \caption{\CausalMatte}
         \label{fig:inpainting_CM_input}
     \end{subfigure}

\par\medskip 
     \begin{subfigure}[b]{0.113\textwidth}
         \centering
         \includegraphics[width=1\textwidth]{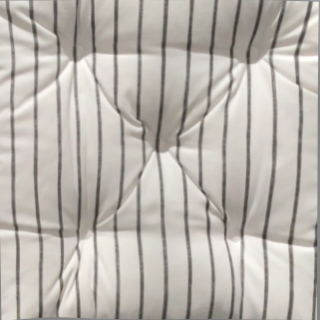}
         \vspace{-1.3\baselineskip}
         \caption{Static Cushion}
         \label{fig:inpainting_static}
     \end{subfigure}
     \begin{subfigure}[b]{0.113\textwidth}
         \centering
         \includegraphics[width=1\textwidth]{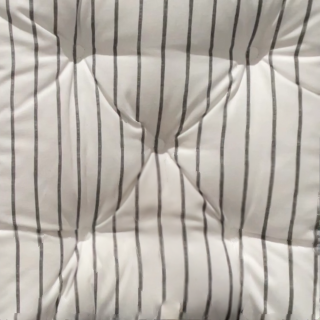}
         \vspace{-1.3\baselineskip}
         \caption{Result w/ (b)}
         \label{fig:inpainting_OM}
     \end{subfigure}
     \begin{subfigure}[b]{0.113\textwidth}
         \centering
         \includegraphics[width=1\textwidth]{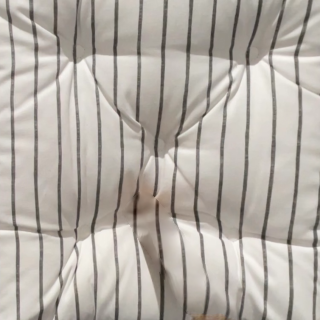}
         \vspace{-1.3\baselineskip}
         \caption{Result w/ (c)}
         \label{fig:inpainting_orig}
     \end{subfigure}
     \begin{subfigure}[b]{0.113\textwidth}
         \centering
         \includegraphics[width=1\textwidth]{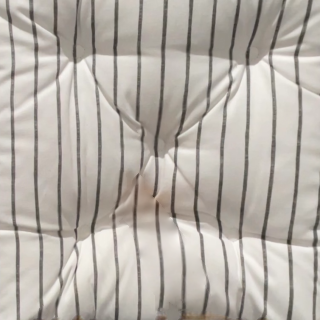}
         \vspace{-1.3\baselineskip}
         \caption{Result w/ (d)}
         \label{fig:inpainting_CM}
     \end{subfigure}
     \caption{\textbf{Inpainting.} We evaluate different methods on the inpainting task of Flow-Guided Video Completion~\shortcite{gao2020flow}, using the mask shown in blue. For \OmniMatte (\subref{fig:inpainting_OM_input}) and \CausalMatte (\subref{fig:inpainting_CM_input}), the masks are acquired by binarizing the predicted alpha channel, using the same threshold for the two methods. (\subref{fig:inpainting_seg_input}) uses the original segmentation mask. For easier comparison, we show an input frame of the static cushion without the stick in (\subref{fig:inpainting_static}). The result (\subref{fig:inpainting_CM}) using \CausalMatte alpha successfully removes the shadow while preserving the deformation near the bottom of the cushion.}
     \label{fig:inpainting}
\end{figure}

The output of \CausalMatte can also be combined with other methods for downstream applications. One good example is object removal. While the counterfactual background video produced by \CausalMatte already is somewhat akin to the result of object removal, inpainting the areas behind removed objects is not our primary goal, and it is done based only on information from the provided video. On the other hand, state-of-the-art texture inpainting methods are specialized for this task and can draw from large datasets for training. Such methods require an input mask to indicate the region to be inpainted, and therefore the quality of this mask can greatly influence the results.
In Figure \ref{fig:inpainting}, we compare the results of Flow-edge Guided Video Completion (FGVC)~\cite{gao2020flow} (a flow-based video completion algorithm) using different input masks provided by a variety of matting methods. 
Simple segmentation masks tend to leave correlated effects like shadows in the scene (Figure \ref{fig:inpainting_orig}), while masks from background subtraction--based methods lead to removal of most interaction effects, including deformations of the background (Figure \ref{fig:inpainting_OM}). The mask from \OurMethod contains the object and its shadow but not the cushion, thus leading to the most plausible invisible result as shown in Figure \ref{fig:inpainting_CM}. 
\begin{figure}[t]
     \centering
     \begin{subfigure}[b]{0.15\textwidth}
         \centering
         \includegraphics[width=1\textwidth]{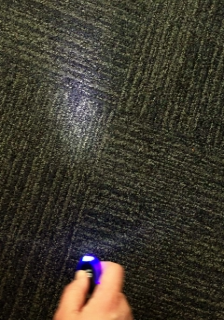}
         \caption{Input Frame}
         \label{fig:color_pop_gt}
     \end{subfigure}
     \begin{subfigure}[b]{0.15\textwidth}
         \centering
         \includegraphics[width=1\textwidth]{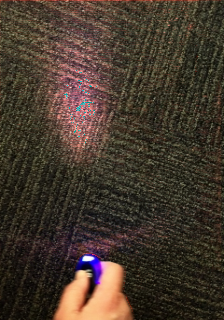}
         \caption{\CausalMatte}
         \label{fig:color_pop_CM}
     \end{subfigure}
          \begin{subfigure}[b]{0.15\textwidth}
         \centering
         \includegraphics[width=1\textwidth]{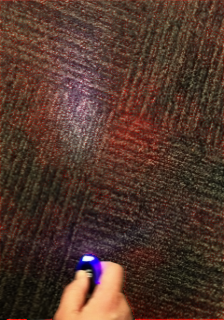}
         \caption{\OmniMatte}
         \label{fig:color_pop_OM}
     \end{subfigure}
     \caption{\textbf{Color Pop.} We can apply color editing to the foreground layer. While \CausalMatte successfully changes the color of the light cast by the flashlight to red, \OmniMatte fails to produce the desired result.}
     \label{fig:color_pop}
\end{figure}

Improving the quality of alpha mattes also enables us to shift the color or timing of components within a video more aggressively than previous methods. The video in Figure \ref{fig:color_pop} shows a person waving a flashlight over a textured surface. Here, we successfully change the color of the flashlight's beam by adjusting the foreground color layer to be more red outside of the input foreground mask.

\begin{figure}[t]
     \centering
     \begin{subfigure}[b]{0.23\textwidth}
         \centering
         \includegraphics[width=1\textwidth]{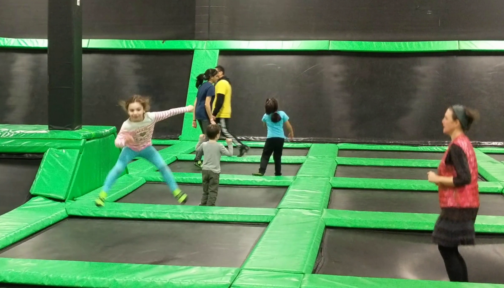}
         \caption{Original Frame}
         \label{fig:retiming_before}
     \end{subfigure}
     \begin{subfigure}[b]{0.23\textwidth}
         \centering
         \includegraphics[width=1\textwidth]{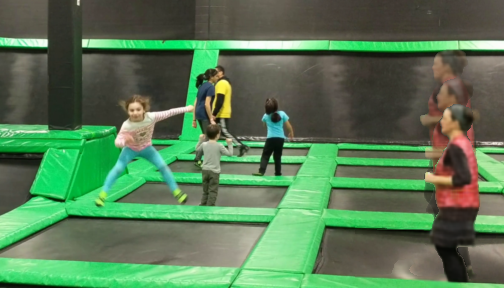}
         \caption{Edited Frame}
         \label{fig:retiming_after}
     \end{subfigure}
     \caption{\textbf{Freezing Time.} For the figure on the right, we composite together several different frames showing them in the act of jumping on the trampoline, while freezing the other smaller figures on the left of the frame.}
     \label{fig:retiming}
\end{figure}

Re-timing tasks are more forgiving of matting errors because mistakes in the alpha channel of static scene content will not typically lead to major artifacts. Therefore, prior methods can already produce high-quality results for re-timing applications, and \CausalMatte performs equally well as shown in Figure \ref{fig:retiming}.

\subsection{Ablations}
\label{subsec:ablation}
\begin{figure*}[t]
    \centering
    \captionsetup[subfloat]{labelformat=parens}
     \begin{subfigure}[b]{0.15\textwidth}
         \centering
         \includegraphics[width=\textwidth]{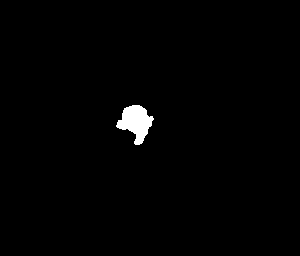}
     \end{subfigure}
     \begin{subfigure}[b]{0.15\textwidth}
         \centering
         \includegraphics[width=\textwidth]{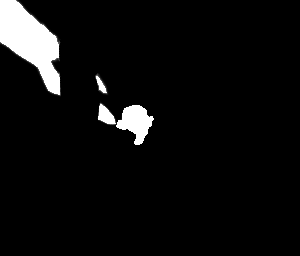}
     \end{subfigure}
     \begin{subfigure}[b]{0.15\textwidth}
         \includegraphics[width=\textwidth]{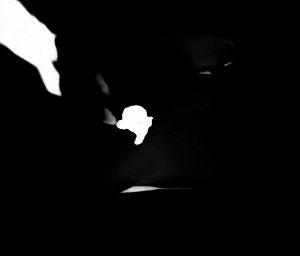}
     \end{subfigure}
     \begin{subfigure}[b]{0.15\textwidth}
         \centering
         \includegraphics[width=\textwidth]{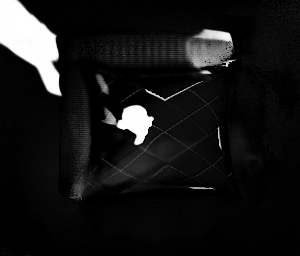}
     \end{subfigure}
     \begin{subfigure}[b]{0.15\textwidth}
         \centering
         \includegraphics[width=\textwidth]{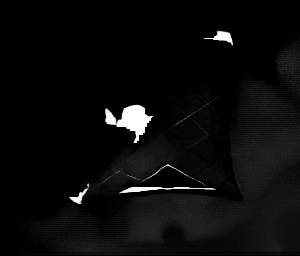}
     \end{subfigure}
     \begin{subfigure}[b]{0.15\textwidth}
         \centering
         \includegraphics[width=\textwidth]{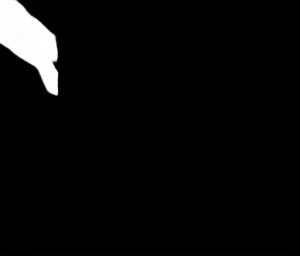}
     \end{subfigure}
     
     \begin{subfigure}[b]{0.15\textwidth}
         \centering
         \includegraphics[width=\textwidth]{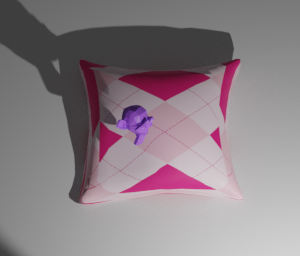}
         \caption{Inputs}
         \label{fig:ablation_inputs}
     \end{subfigure}
     \begin{subfigure}[b]{0.15\textwidth}
         \centering
         \includegraphics[width=\textwidth]{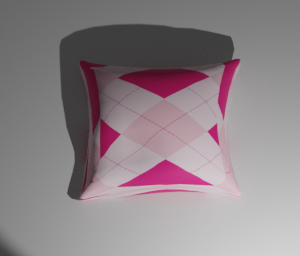}
         \caption{Ground Truth}
         \label{fig:ablation_gt}
     \end{subfigure}
     \begin{subfigure}[b]{0.15\textwidth}
         \centering
         \includegraphics[width=\textwidth]{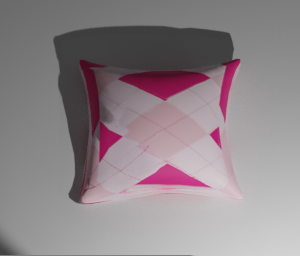}
         \caption{Full Method}
         \label{fig:ablation_full_method}
     \end{subfigure}
     \begin{subfigure}[b]{0.15\textwidth}
         \centering
         \includegraphics[width=\textwidth]{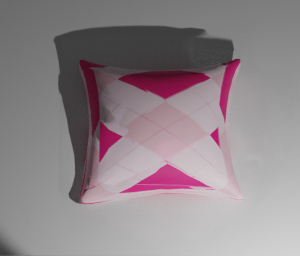}
         \caption{w/o \tLest}
         \label{fig:ablation_flow0}
     \end{subfigure}
     \begin{subfigure}[b]{0.15\textwidth}
         \centering
         \includegraphics[width=\textwidth]{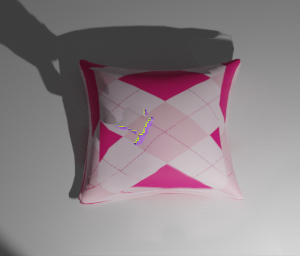}
         \caption{w/o \tLAwarp}
         \label{fig:ablation_step1}
     \end{subfigure}
     \begin{subfigure}[b]{0.15\textwidth}
         \centering
         \includegraphics[width=\textwidth]{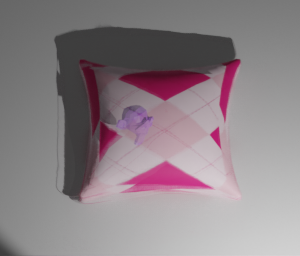}
         \caption{w/o Residual $\Discr{}$}
         \label{fig:ablation_dis03}
     \end{subfigure}
     \caption{\textbf{Ablations.} (\subref{fig:ablation_inputs}) shows the input mask (top) and frame (bottom) for a frame from the \textsc{Purple Monkey} video. For the rest of this figure, the first row shows the \FL's alpha channel for various ablation experiments, and the second row shows the counterfactual background component generated by compositing the \RL and the \BL.  (\subref{fig:ablation_gt}) Simulated ground truth; (\subref{fig:ablation_full_method}) our full method; (\subref{fig:ablation_flow0}) results when turning off the flow estimate loss; (\subref{fig:ablation_step1}) results when turning off consistency loss; (\subref{fig:ablation_dis03}) results when turning off the \RL discriminator. With pixel values ranging from -1 to 1, the average per-pixel L1 errors of the \FL alpha with respect to ground truth (\subref{fig:ablation_gt}) for (\subref{fig:ablation_full_method}), (\subref{fig:ablation_flow0}), (\subref{fig:ablation_step1}) and (\subref{fig:ablation_dis03}) respectively are 0.007, 0.027, 0.027, and 0.012.}
     \label{fig:ablation}
\end{figure*}

In Figure \ref{fig:ablation} we ablate our method by removing individual losses. 
We use a synthetic video to facilitate comparisons with ground truth counterfactuals.
We find that when \tLest is removed (Figure \ref{fig:ablation_flow0}), the decomposition fails to group some pixels that are primarily related through their correlated flow. With only color information, the network is more prone to mistakes that are characteristic of image-only matting. This is particularly problematic if the distributions of color in different layers overlap. In our test video, the cushion and the monkey cast shadows of similar colors, causing this ablated network to mistakenly assigns parts of the cushion shadow to the \FL. 

Learning the correct flow alone is not sufficient for videos with complex interactions. In Figure \ref{fig:ablation_step1} the alpha consistency regularization is turned off, and we find that the \FL alpha flickers across frames. Without the coherence constraint, correct predictions before and after interactions no longer help as much with the more difficult factorization of frames with heavy interaction effects. 

If there is complex layer that lacks a discriminator, then the model can assign spurious contents to that layer rejected by other discriminators. In Figure \ref{fig:ablation_dis03}, we turn off the discriminator for \RL. Since a part of the monkey is refused by the \FL discriminator, the background counterfactual component contains it as an artifact.

\section{Limitations and Future Work}
\label{sec:discussion}

\CausalMatte works effectively when the conditional appearance of different components is well-separated---in other words, when the distributions represented by our conditional priors are easy to differentiate. It becomes more difficult when different components have more similar appearance. In such cases, balancing the weights of different losses in our framework can be a delicate task. To some extent this limitation is fundamental: if the foreground and background look identical then there is little hope to separate them. Pre-training priors with much larger datasets could allow us to represent those priors more accurately, potentially leading to more separated distributions.

For very complex cross-component interactions it can be difficult to find \interactionprior{}s that span the conditional appearance of each component. If the \posex used to train each conditional prior are not close enough to the ideal counterfactual distributions, the input video may appear conditionally improbable to all components. In this case, the assignment of content to layers becomes more arbitrary and will often be wrong. The \textsc{Puddle Splash} results, despite being impressive in context, still suffer some visible artifacts for this reason. We found that turning down the \FL discriminator loss for 100 epochs at the end of training reduced these artifacts (The original results before this step can be found in the Appendix). Empirically this method improves other videos with similar issues, which suggests that despite the effectiveness of the current version of our method, there are still performance gains to be achieved through parameter tuning.

The most significant practical limitation of our current implementation is runtime. As our method contains multiple training stages and additional networks, our average runtime is approximately 3.5 times that of \OmniMatte. 
In this work, we primarily focused on developing the factor matting problem formulation and a general framework for solving it, and left exploring more efficient variants to the future.

\section{Conclusion}
\label{sec:conclusion}
We have reframed video matting in terms of counterfactual video synthesis for downstream re-compositing tasks, where each counterfactual video answers a questions of the form ``what would this component look like if we froze time and separated it from the rest of the scene?’’
We connected this definition to a Bayesian formulation of the problem where the typical assumption of independent layers has been removed, and we presented a framework for learning conditional priors by applying a distribution of transformations to samples from marginal priors. 
Finally, we showed that this strategy leads to cleaner video mattes and more useful factoring of color into different video layers, and we demonstrated how to use the decompositions produced by our method in various editing applications.

We have demonstrated that \OurMethod outperforms existing state-of-the-art methods on many tasks and that our results can benefit a wide range of downstream applications. We believe this will offer a powerful tool for video editing and analysis and hope that our work will inspire further exploration in this direction.

\bibliographystyle{ACM-Reference-Format}

\section{Appendix}
\label{sec:supp}
\subsection{Training Details}
\subsubsection{Alpha Initialization}
Our decomposition network \tdecompN takes inputs in a similar format to \OmniMatte~\shortcite{lu2021omnimatte}: for each layer at time step t, the inputs are the reference mask, estimated flow in the mask region, and a sampled noise image $Z_t$ representing the background. Here, we distinguish \emph{reference masks} from the \emph{segmentation (input) masks} described in the main paper. 
By segmentation or input mask, we mean the raw user input; these are are automatically pre-processed to become a set of reference masks that are the actual inputs to the system. Below, we describe how this pre-processing works.

The reference mask for each layer should be distinct to help the network to distinguish between layers. For the \FL, we build from the segmentation mask. 
For the \RL, since it is designed to describe the entire background, its reference mask should cover the entire frame.
At the same time, we observe that at the beginning of training, the decomposition network is biased towards predicting similar alpha values for positions with the same reference mask value. 
Therefore, if we directly use the binary input segmentation as the reference mask for \FL, and a constant $\mathbf{1}$ as the reference mask for \RL, then for regions outside the input segmentation mask, the \RL reference mask would equal $\mathbf{1}$, whereas the \FL mask would equal $\mathbf{0}$. 
This setting would bias the model to attribute those pixels entirely to the \RL, rather than producing the desired split into a composition of factors. 

Instead, we reason as follows. For an opaque foreground object, its input segmentation mask encodes the user's belief that the appearance of the input video at those masked pixels should be attributed entirely to the \FL. 
In contrast, for a transparent object, or for regions outside the mask, any layer might contribute to the final appearance. 
Thus, taking a probabilistic view, in regions covered by the original segmentation mask, we set the reference masks of the \FL to \textbf{1}, and those of the \RL to \textbf{0}. Outside the masked region, we set the reference mask value uniformly to \textbf{0.5} for both the \FL and \RL to express the uncertain attribution into layers. The input flow for \FL is masked by the original binary segmentation, and for \RL it is the entire frame. The \BL is constrained to be a fixed opaque layer transformed by a per-frame homography, and thus is not influenced by the reference mask.

\subsubsection{Alpha Regularization}
We adopt two regularizers on the predicted alpha channels from \OmniMatte. \tLsparsity uses a mixture of an $L_1$ and an approximate $L_0$ loss to encourage sparsity: 
\begin{equation}
    \Lsparsity = \frac{1}{T}\frac{1}{N}\sum_{t}\sum_{i}\gamma||\LayerAlpha{i}(t)||_1 +\Phi_0(\LayerAlpha{i}(t))
\end{equation}
where $\Phi_0(x) = 2 \cdot Sigmoid(5x)-1$, and $\gamma$ controls the relative weighting of the two terms. \tLmask expedites convergence by forcing the initial \FL (index 1) alpha channel to be close to the segmentation masks $M_1(t)$.
\begin{equation}
    \Lmask = \frac{1}{T}\sum_{t}||d_1(t) \odot (M_1(t)-\LayerAlpha{1}(t))||_1
\end{equation}
$d_1(t)=1-\textit{dilate}(M_1(t))+M_1(t)$ gradually turns off the loss near the mask boundary such that the model is less sensitive to small noise and errors, and $\odot$ is the element-wise product. This loss is turned off after a fixed number of epochs or if the loss is lower than a threshold.

\subsubsection{Training Parameters}
We feed random noise to initialize the decomposition network in the same way as Omnimatte~\shortcite{lu2021omnimatte}. 
We set up one discriminator for the foreground layer, and one for the residual layer. 
Their receptive fields are at three scales: 16$\times$16, 32$\times$32, and 64$\times$64. The patches are sampled over the entire frame, so that the discriminator loss \tLadv provides an alternative supervision to zero-alpha regions that cannot be supervised by \tLrecon (\tLCwarp and \tLAwarp are other supervisions on those regions). In Stage 1, we use SIFT features~\cite{lowe2004distinctive} and FLANN-based matcher to pre-compute a homography for each frame. Thus the model only needs to estimate a background canvas and warp it to each frame with an appropriate homography. Then we freeze the \BL in the later stages to simplify the factorization optimization.
We use all aforementioned losses and train on NVIDIA RTX A6000 for 1200 epochs in both Stage 1 and 3, and only use \tLadv in Stage 3.
We use the ADAM optimizer~\cite{kingma2014adam} when training the decomposition network and discriminators, with learning rate 0.001 and batch size 16. When applying strategies in Section \ref{sec:userinputs}, we use 3 temporal scales: the frame at time step $t$ is paired with $t+1$, $t+4$, and $t+8$ as inputs. All flow-related losses are optimized for all scales.

Assuming that we have two discriminators $\{\Discr{1}, \Discr{2}\}$: one for \FL and one for \RL, then the complete loss for each of them and for the decomposition network \tdecompN is:
\begin{equation}
\begin{split}
    L_{N_D} &= 
    \Lrecon
    +\lambda_1 \LCwarp
    +\lambda_2 \LAwarp\\
    & +\lambda_3(\Ladv^1+\Ladv^2)
    +\lambda_4\Lsparsity
    +\lambda_5\Lest\\
    & +\lambda_6\Lmask
\end{split}
\end{equation}
\begin{equation}
L_{\Discr{i}} = \lambda_7\Ladv^i
\end{equation}
where $\lambda_1 = \lambda_2 = \lambda_3 = 0.005$, $\lambda_4 = \lambda_5 = \lambda_7 = 0.0005$, $\lambda_6 = 25$, and \tLmask is turned off after it falls below 0.05.

\subsubsection{Model Architecture}
Our decomposition network architecture, described in the table below, is similar to that of \cite{lu2020layered}. Layers 13a and 13b are two parallel convolution blocks: the first produces the final RGB prediction, and the second produces alpha and two-dimensional flow.
\begin{center}
\begin{tabular}{ |c|c|c|c|c| } 
\hline
{} & Layer type(s) & channels & stride & activation\\
\hline
1 & conv & 64 & 2 & leaky \\ 
2 & conv, BN & 128 & 2 & leaky \\ 
3 & conv, BN & 256 & 2 & leaky \\ 
4 & conv, BN & 256 & 2 & leaky \\ 
5 & conv, BN & 256 & 2 & leaky \\ 
6 & conv, BN & 256 & 1 & leaky \\ 
7 & conv, BN & 256 & 1 & leaky \\ 
8 & skip5, convt, BN & 256 & 2 & relu \\ 
9 & skip4, convt, BN & 256 & 2 & relu \\ 
10 & skip3, convt, BN & 128 & 2 & relu \\ 
11 & skip2, convt, BN & 64 & 2 & relu \\ 
12 & skip1, convt, BN & 64 & 2 & relu \\ 
13a & conv & 3 & 1 & tanh \\ 
13b & conv & 3 & 1 & none \\ 
\hline
\end{tabular}
\end{center}
The architecture of our discriminators is similar to PatchGAN. We set kernel size to 4 for all layers as below:
\begin{center}
\begin{tabular}{ |c|c|c|c|c| } 
\hline
{} & Layer type(s) & channels & stride & activation\\
\hline
1 & conv & 64 & 2 & leaky \\ 
2 & conv, BN & 128 & 1 & leaky \\ 
3 & conv & 1 & 1 & none \\ 
\hline
\end{tabular}
\end{center}

\subsection{Additional Results}
\subsubsection{\textsc{Puddle Splash} Before Post-Processing}
\begin{figure}[t]
     \centering
     \begin{subfigure}[h]{0.113\textwidth}
         \centering
         \includegraphics[width=\textwidth]{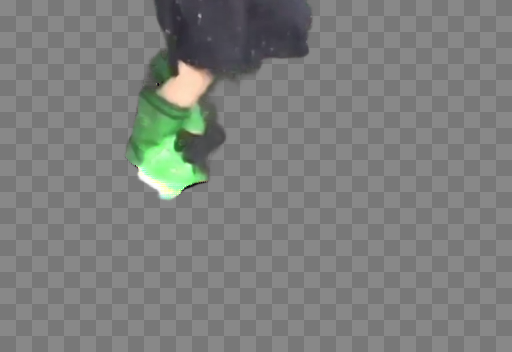}
         \caption{Foreground \protect\\ Layer RGBA}
         \label{fig:supp_splash_rgba_l2}
     \end{subfigure}
     \begin{subfigure}[h]{0.113\textwidth}
         \centering
         \includegraphics[width=\textwidth]{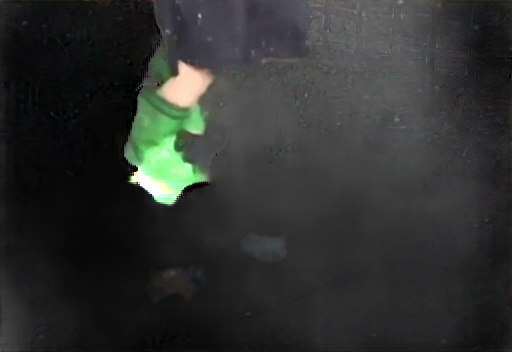}
         \caption{Foreground \protect\\ Layer RGB}
         \label{fig:supp_splash_rgb_l2}
     \end{subfigure}
          \begin{subfigure}[h]{0.113\textwidth}
         \centering
         \includegraphics[width=\textwidth]{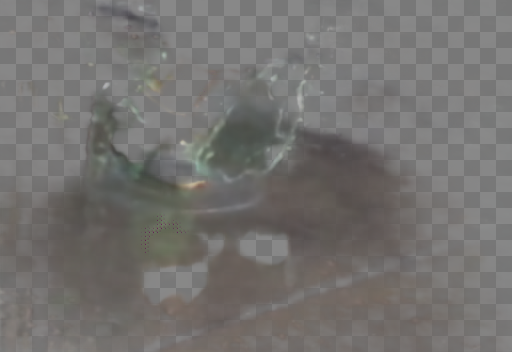}
         \caption{Residual \protect\\ LayerRGBA}
         \label{fig:supp_splash_rgba_l1}
     \end{subfigure}
     \begin{subfigure}[h]{0.113\textwidth}
         \centering
         \includegraphics[width=\textwidth]{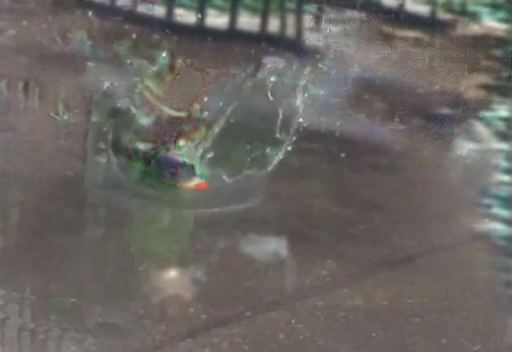}
         \caption{Residual \protect\\ Layer RGB}
         \label{fig:supp_splash_rgb_l3}
     \end{subfigure}
     \caption{\textbf{\textsc{Puddle Splash} Before Post-Processing.}}
\end{figure}
The foreground and background components in our teaser video are particularly entangled, due to the complex reflections between the foot and the puddle, whose color space our default \posex generation for \FL failed to span. 
Therefore, our \FL discriminator became overly restrictive and \tLrecon forced some of the foot reflection into the more inclusive \RL discriminator. 
If we tune down weights on the \tLrecon and \FL discriminator a bit, then the \RL discriminator would exert its power in full, and give the results in Figure \ref{fig:teaser}.

\subsubsection{Ablation on Discriminator Patch Size}
\begin{figure}[t]
     \centering
     \begin{subfigure}[h]{0.23\textwidth}
         \centering
         \includegraphics[width=\textwidth]{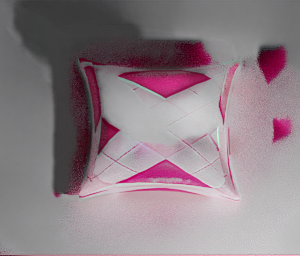}
         \caption{Residual RGB w/ 7$\times$7 Patches}
         \label{fig:supp_patchsize7}
     \end{subfigure}
     \begin{subfigure}[h]{0.23\textwidth}
         \centering
         \includegraphics[width=\textwidth]{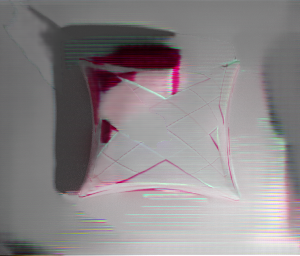}
         \caption{Residual RGB w/ 70$\times$70 Patches}
         \label{fig:supp_patchsize70}
     \end{subfigure}
  \caption{\textbf{Ablation on Patch Size.} We show the \RL RGB w/ single-scale discriminators of different patch size. The video is the one used in Figure \ref{fig:ablation}. In (\subref{fig:supp_patchsize7}) we use a patch size of $7\times7$, and in (\subref{fig:supp_patchsize70}) it is $70\times70$.}
  \label{fig:supp_patchsize}
\end{figure}

The input patch size influences how much global structure information the discriminators can see. In Figure \ref{fig:supp_patchsize}, we compare the \RL color layer from a discriminator with input patch size $7\times7$ (Figure \ref{fig:supp_patchsize7}) against one with $70\times70$ (Figure \ref{fig:supp_patchsize70}). When the patch size is small, the \RL discriminator is relaxed on global structure and allows the pink pattern to overflow to outside the cushion. It also allows the foreground object's shadow due to its similar color with the cushion's shadow. A larger patch size alleviates such mistakes.

\end{document}